%% file: acl_latex.tex
\pdfoutput=1

\documentclass[11pt]{article}

\usepackage[final]{acl}

\usepackage{times}
\usepackage{latexsym}

\usepackage[T1]{fontenc}

\usepackage[utf8]{inputenc}

\usepackage{microtype}

\usepackage{inconsolata}

\usepackage{graphicx}

\usepackage{algorithm}
\usepackage{algorithmic}
\usepackage{tabularx}
\usepackage{booktabs}
\usepackage{colortbl}
\usepackage{multirow}
\usepackage{amsmath}

\title{Consistent Client Simulation for Motivational Interviewing-based Counseling}

\author{
 \textbf{Yizhe Yang\textsuperscript{1}~\thanks{Work was done during a visit at SMU.}},
 \textbf{Palakorn Achananuparp\textsuperscript{2}},
 \textbf{Heyan Huang\textsuperscript{1}~\thanks{Corresponding Author}},
 \textbf{Jing Jiang\textsuperscript{3}},
\\
 \textbf{John Pinto\textsuperscript{4}},
 \textbf{Jenny Giam\textsuperscript{5}},
 \textbf{Kit Phey Leng \textsuperscript{6}},
 \textbf{Nicholas Gabriel Lim \textsuperscript{7}},
 \\
 \textbf{Cameron Tan Shi Ern \textsuperscript{8}},
 \textbf{Ee-peng Lim\textsuperscript{2}}
\\
\\
 \textsuperscript{1}Beijing Institute of Technology,
 \textsuperscript{2}Singapore Management University,
 \textsuperscript{3}Australian National University, 
 \\
 \textsuperscript{4}ThoughtFull,
 \textsuperscript{5}Singapore Institute of Technology,
 \textsuperscript{6}National Institute of Education,
 \\
 \textsuperscript{7}Singapore University of Social Sciences,
 \textsuperscript{8}National University of Singapore
\\
 \small{
   \{yizheyang,hhy63\}@bit.edu.cn, \{palakorna,eplim\}@smu.edu.sg, jing.jiang@anu.edu.au
 }
}

\begin{document}
\maketitle
\begin{abstract}
Simulating human clients in mental health counseling is crucial for training and evaluating counselors (both human or simulated) in a scalable manner. Nevertheless, past research on client simulation did not focus on complex conversation tasks such as mental health counseling. In these tasks, the challenge is to ensure that the client's actions (i.e., interactions with the counselor) are consistent with with its stipulated profiles and negative behavior settings. In this paper, we propose a novel framework that supports consistent client simulation for mental health counseling. Our framework tracks the mental state of a simulated client, controls its state transitions, and generates for each state behaviors consistent with the client's motivation, beliefs, preferred plan to change, and receptivity.  By varying the client profile and receptivity, we demonstrate that consistent simulated clients for different counseling scenarios can be effectively created.  Both our automatic and expert evaluations on the generated counseling sessions also show that our client simulation method achieves higher consistency than previous methods.
\end{abstract}

\input{sections/introduction}

\input{sections/related_work}

\input{sections/client_simulation_framework}

\input{sections/experiment}

\section{Conclusion}
In this paper, we focus on client simulation for Motivational Interviewing (MI) counseling, a complex scenario where the client should exhibit varied behaviors in different counseling stages. We introduce a consistent client simulation framework that integrates state tracking and action selection with an LLM to accurately mimic the behaviors of real clients according to given profiles. Based on our framework, we develop a client simulation method that supports changes of client states consistent with real world MI-counseling using state transition control, and utterances consistent with real world client's profile using action and information selection modules. These modules utilizes action distribution from real datasets for different levels of receptivity. The action selection mechanism incorporates both context-aware and (state,receptivity)-aware action distributions, to align with the expected client behavior. Extensive evaluations using GPT-4 and human assessments demonstrate the high consistency of our method with real clients, outperforming the existing client simulation methods. With our client simulation framework, we believe that more realistic simulated clients can be implemented for training and evaluating counselors.

\section*{Limitations}
Although our simulation framework shows potential for future use in counseling, there are three main limitations to this work. 

Similar to other LLM prompting approaches, our framework also suffers from prompt sensitivity, as it incorporates multiple steps of prompting, which may lead to error accumulation. To mitigate such sensitivity, we employ a few-shot method to formulate our prompts. The design of effective prompts and their evaluation are therefore important areas for further research.

This paper focuses on client simulation only.  As a good counseling session also depends on the performance of counselor, we also need to develop highly competent counselor agents and evaluate them. Nevertheless, without a diverse set of clients, it is very difficult to meet this objective. In our future work, we will therefore leverage our client simulation framework to design better counselor agents and to develop a robust approach to evaluate their counseling skills.

Our current study focuses on real client simulation and conduct experiments on a single dataset (i.e. AnnoMI). We also restrict our study to MI approach to counseling.  Our work however can be easily extended to involve more real world counseling datasets as long as privacy and ethical concerns are addressed. We can also extend our framework to accommodate other counseling approaches (e.g., Cognitive behavioral therapy) as it is quite common for experienced counselors to apply a mixture of different approaches in a single counseling session so that to both personalize to the real clients and achieve the best outcome. 

\section*{Ethics Statement}
For the revision of the framework and any related annotation work, we conducted a thorough survey of psychological knowledge and datasets pertaining to motivational interviewing counseling. All annotations were performed by our co-authors.


\bibliography{custom}

\appendix

\input{sections/data_annotation}

\input{sections/human_annotation}

\input{sections/more_evaluation}

\input{sections/implement}

\input{sections/case}

\end{document}

%% file: sections/introduction.tex
\section{Introduction}

Traditionally, mental health counselors are trained with human clients. To address the cost and limited access to human clients, researchers begin to explore the use of large language models (LLMs) to create agent-simulated clients~\citep{yosef2024assessing, chiu2024computational, wang2024towards}. Effective mental health counseling requires counselors to apply some principled approach tailored to the given client's problem. Motivational Interviewing (MI) is one such approach that aims to elicit client-driven motivation towards behavioral change~\citep{miller2012motivational,prochaska1997transtheoretical, hashemzadeh2019transtheoretical}. In MI, a counselor aims to guide a client in the Precontemplation state denying, and downplaying his/her behavioral problem towards the Preparation state where the client starts to plan for behavioral change~\citep{abuse2019enhancing}. To train a counselor who can adapt to the unique nuances of many different types of clients\citep{hoang2024can}, we require agent-simulated clients to closely mimic human clients of different profiles and behaviors in highly interactive counseling sessions.

\begin{figure}[tb]
    \centering
    \includegraphics[width=0.49\textwidth]{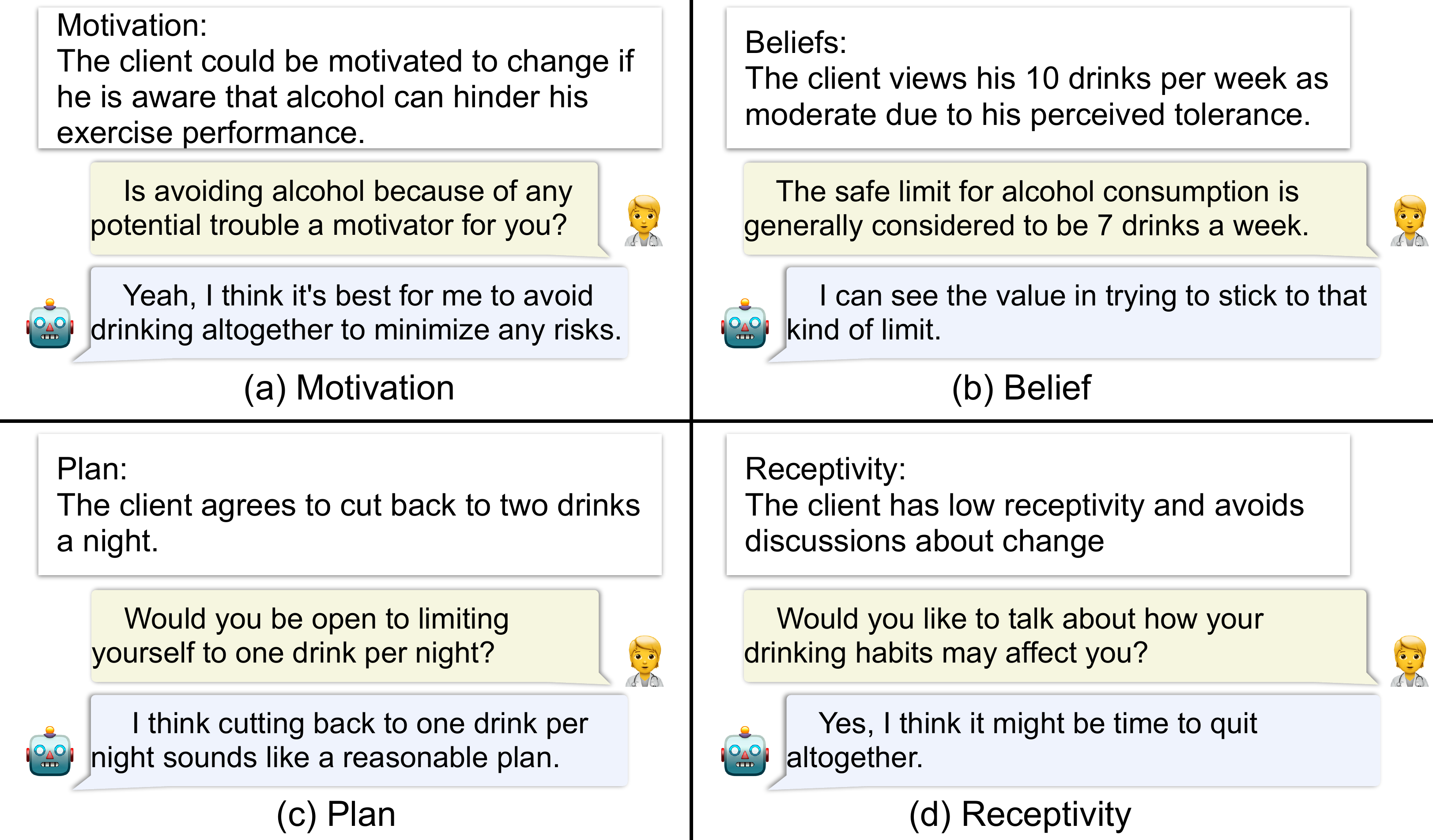}
    \caption{Types of inconsistency in existing client simulation approaches (Left: simulated client, Right: counselor):  client response inconsistent with the (a) client's motivation to change,  (b) beliefs, (c) preferred plans to change, and (d) receptivity.}
    \label{fig:inconsistency}
\end{figure}

Nevertheless, the current client simulation methods predominantly utilize simple personas~\citep{yosef2024assessing,wang2024towards} or conversation examples of the clients~\citep{chiu2024computational} as LLM prompts. These prompts focus on personas but ignore the diverse dialogue actions and nuanced state transitions during counseling, which in turn leads to generated utterances inconsistent with client profiles and behaviors. Moreover, LLMs like ChatGPT are aligned to generate compliant responses~\citep{shen2023large, kopf2024openassistant}. Simulated clients using simple LLM prompting are found to demonstrate overly-compliant behavior or select overly-narrow set of actions compared to human clients~\citep{kang2024can}. 

Based on our empirical evaluation of previous client simulation approaches~\citep{yosef2024assessing,chiu2024computational,wang2024towards}, we determine four types of inconsistencies as depicted in Figure~\ref{fig:inconsistency}. The simulated clients are likely agree to behavior change for reasons not aligned with their input motivation or accept plans inconsistent with the preferred plans as shown in Figures~\ref{fig:inconsistency}(a) and  (c) respectively. Figure~\ref{fig:inconsistency}(b) also shows a simulated client not adhering to the given beliefs. Moreover, the simulated client may fail to demonstrate the stipulated receptivity to others' opinions as shown in Figure~\ref{fig:inconsistency}(d).

To tackle these challenges, we develop a comprehensive framework for consistent client simulation, as illustrated in Figure~\ref{fig:framework}. Our framework comprises four key modules: state transition, action selection, information selection, and response generation modules. Unlike the previous approaches, our framework explicitly models four aspects of client profiles, i.e., motivation, beliefs, preferred change plans, and receptivity. The framework's modules jointly control the simulated client's state and behavior at a fine-grained level thereby achieving a high degree of consistency with respect to input client profile and MI-counseling process. To achieve high degree of realism, the modules utilize knowledge extracted from a real world MI-counseling dataset.

We summarize our contributions as follows:

\begin{itemize}
    \item Our work is the first to define different aspects of consistency. We incorporate state tracking, action selection, and information selection in our proposed client simulation framework to  jointly support the overall consistency of a simulated client's utterances with respect to his/her profile and expected behavior in MI-counseling.
    \item By utilizing action distribution and state transition knowledge of clients derived from a real world counseling dataset, our client simulation framework can be easily adapted to different counseling use cases where relevant real world datasets are available. 
    \item An extensive set of experiments involving both automatic and expert evaluation shows that our client simulation method achieves higher consistency and creates more human-like simulated clients than other baseline methods.
\end{itemize}

%% file: sections/related_work.tex
\section{Related Work}

\paragraph{Motivational Interviewing}
Motivational Interviewing (MI) is a client-centered counseling approach for eliciting behavior change by helping clients to explore and resolve ambivalence~\citep{miller2012motivational}. Recent works on automated MI-counselling ~\citep{chiu2024computational,yosef2024assessing,wang2024towards} predominantly focus on the evaluation of counselor agents, but ignoring the simulation of clients with strong consistency. As these counselor agents interact with hypothetical clients, their effectiveness in the simulated MI-counselling sessions would not accurately reflect that involving real human clients. To the best of our knowledge, our work is the first to address the consistency of client simulations in MI counseling. MI outlines a progression through five stages of change: precontemplation, contemplation, preparation, action, and maintenance~\citep{prochaska1997transtheoretical,hashemzadeh2019transtheoretical}. In our work, we focus on the states from Precontemplation to Preparation which are most challenging to counselors.

\paragraph{Client Agent Simulation}
Client agent simulation provides a cost-effective way to mimic the behaviors of real users and is extensively employed across different patient simulation applications~\citep{wang2024patient}. \citet{wang2024patient} introduced PATIENT-$\Psi$, integrating cognitive models with LLMs to accurately mimic the communication behaviors of real patients. However, they rely on the GPT-4 to infer patients' behaviors from their personas instead of staying consistent with real human client behaviors in counseling sessions. In our method, we integrate state tracking and action selection to enhance the consistency of the client simulations. \citet{liao2024automatic} introduced a State-Aware Patient Simulator that tracks the action of a doctor before selecting specific actions and relevant information to disclose. While their state tracking and action selection approach is similar to ours, we differ by learning the client's state transitions and actions from real datasets and adopting turn-by-turn response generation. Furthermore, patient simulators are designed to accurately disclose information, whereas counseling clients may be reticent or exhibit non-collaborative behaviors. Thus, we incorporate receptivity controls to simulate such realistic interaction.

%% file: sections/client_simulation_framework.tex
\section{Client Simulation Framework}
\label{sec:framework}

\begin{figure*}[tb]
    \centering
    \includegraphics[width=\textwidth]{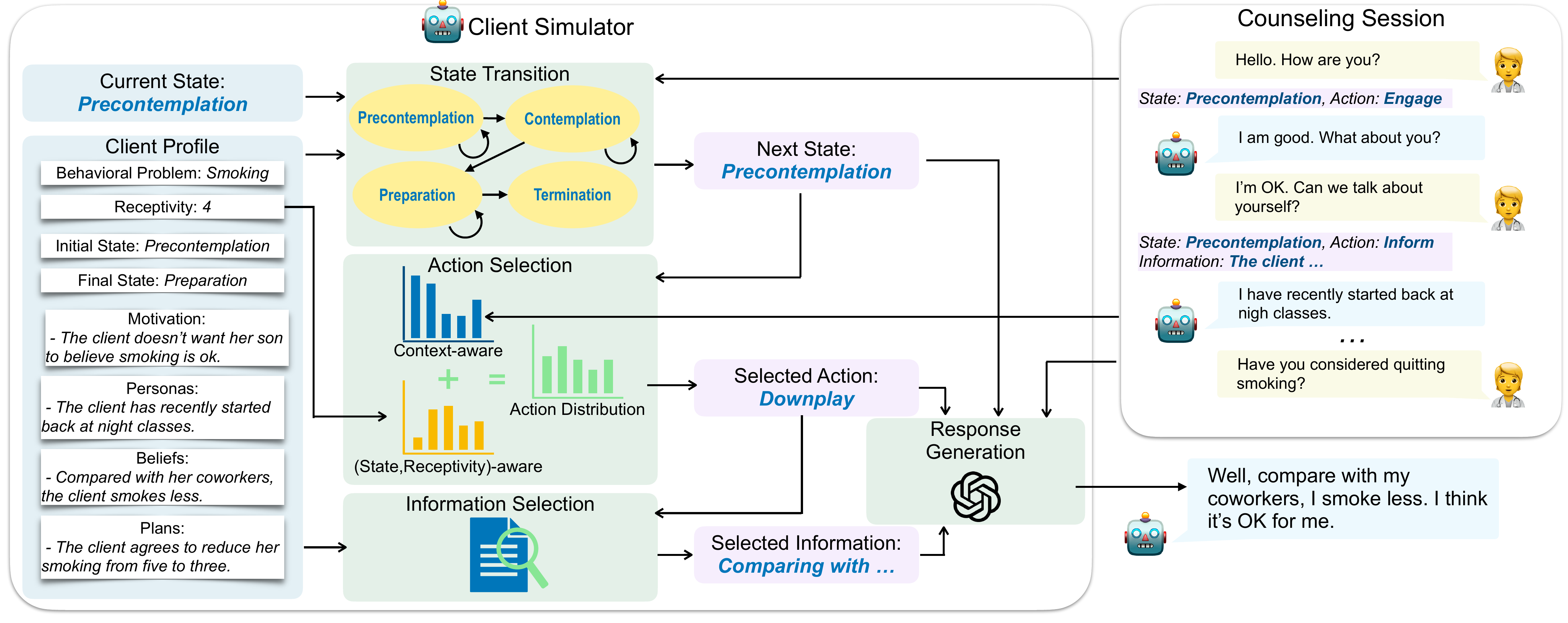}
    \caption{Proposed Client Simulation Framework.}
    \label{fig:framework}
\end{figure*}

We first give an overview of our proposed client simulation framework to generate client utterances consistent to client's profile and behavior in a MI-counseling session.  The framework consists of modules for: (a) {\em state transition}, (b) {\em action selection} (c) {\em information selection} , and (d) {\em response generation} as shown in Figure~\ref{fig:framework}.\footnote{The detailed prompt design of our framework can be found in Appendix~\ref{app:implement}.} Finally, we annotate and analyze the real world MI-based counseling data for deriving the knowledge required for developing modules (a) and (b).

\subsection{Overview}

Our framework takes a {\em client's profile} as input. It consists of the client's behavioral problem, initial state, final state, persona, motivation, beliefs, preferred change plans, and receptivity.  The client's behavioral problem, persona, motivation, beliefs, and preferred change plans are expressed in text. The initial and final states refer to the client's initial and final state-of-mind towards the behavioral problem before and after the counseling session, respectively. We utilize the transtheoretical model of health behavior change ~\citep{abuse2019enhancing} to define three possible states: {\em Precontemplation}, {\em Contemplation}, and {\em Preparation}~\citep{prochaska1997transtheoretical, hashemzadeh2019transtheoretical}. Note that there also exists Action and Maintenance states in the transtheoretical model. As MI-counseling is particularly useful when clients are in the Precontemplation and Contemplation states for them to reach the Preparation state, we consider only the three primary states henceforth, which also aligns well with the real dataset. To end the counseling session, we also include the {\em Termination} state.  

Similar to the earlier work, the persona covers background information about the client. These are useful information the counselor may need to uncover during the session. Motivation refers to a specific reason that can lead the client to consider making changes. In contrast, the client's beliefs are the ones which may obstruct behavioral changes. Preferred change plans (or plan) refer to the set of specific behavior changes the client may agree to. Receptivity captures how receptive the client is towards the counselor and is categorized into five levels from lowest (1) to highest (5). 

The client simulator generates one utterance at a time in the counseling session.  Right after each counselor's utterance, the simulator first determines the client's next state using the state transition module which has the current counseling session text (also known as context) and current state as input.  Conditioned on the next state, the action selection module merges the context-aware and (state, receptivity)-aware action distributions so that an action representing a type of utterance can be sampled.  The sampled (or selected) action, together with the new state and current context, will then be used by the response generation module to generate the next utterance.  If the selected action requires some client's profile knowledge (e.g., Inform, Hesitate, and Blame actions), the simulator will perform an additional information selection step to search for relevant reference information from the profile for response generation.

\subsection{State Transition}
\label{sec:state transition}


During MI counseling, a client is guided through different states to identify changes that can address his/her behavioral problem. The goal of the state transition module is thus to maintain consistency with the way the client may change states during MI counseling as well as the client's profiles~\citep{abuse2019enhancing}.

In the Precontemplation state, the counselor focuses on building trust, establishing change goals, understanding the client's motivations, and address some wrong beliefs, before evoking the client to change. The client is expected to enter the Contemplation state only when motivated by a specific reason which has also been brought up by the counselor. The state transition module thus seeks to be consistent by analyzing the counselor's utterances to search for mention(s) of client's motivation. If such a mention is found, the next client's state is Contemplation; otherwise, the state remains unchanged. In the Contemplation state, the client recognizes his/her behavioral problem but is hesitate to change, often due to some beliefs which are part of client's profile. The state transition module tracks these reasons from the client's profile and assesses whether they have been adequately addressed. If so, the client will transit to the Preparation state. In the Preparation phase, the client discusses specific plans for change while counselor provides the necessary information and tools to guide this discussion. When the client's preferred change plan has been discussed, the state transition module should output Termination as the next state. The state transition in our framework align with the observations in a real dataset. Nevertheless, our proposed framework can accommodate more complex flows and additional states, provided that further comprehensive datasets are available.

\subsection{Action Selection}
\label{sec:action selection}

Inspired by \citet{dutt2021resper}, we tailor client actions to MI counseling and integrate a simple yet effective action sampling method into the action selection module. We meticulously design candidate actions for each state based on MI counseling theory and analysis of a real dataset.\footnote{The detailed candidate actions are outlined in the Appendix~\ref{app:implement}.} Similarly, our proposed framework can accommodate additional actions, provided further datasets are available. We recognize the need to consider both the latest counselling session context, client's state and receptivity in action selection. We first infer the context-aware action distribution based on the latest session context using LLM. We next derive from real world MI-counseling data the action distribution for each (state,receptivity) combination. To ensure both context coherence and profile consistency, we finally merge the context-aware action distribution and (state,receptivity)-aware action distribution by averaging the two.  When sampling from the merged action distribution, we only select among among actions relevant to the next state.

\subsection{Information Selection}
\label{sec:information selection}

Our framework divides actions into two types. The type-1 actions, e.g., Deny, Engage, and Accept, do not require profile information to generate the utterances. The type-2 actions (e.g., Inform, Blame, Hesitate, and Plan) utilize additional information from the client's profile for utterance generation. The information selection module thus selects appropriate information from client profile for type-2 actions using LLM prompting similar to that in \citet{tu2023characterchat}, thereby enhancing the consistency with the client profile. This selection mechanism also prevents the simulated clients from sharing too much profile information unnecessarily shortening the counseling sessions.

\subsection{Response Generation}
\label{sec:response generation}

Finally, the response generation module produces a client's utterance using a {\em turn-by-turn} generation approach. Specifically, using the same application session\footnote{This covers the entire history of prompt instructions, generated counselor and client utterances including the initial role-playing instruction and client profile.} with the LLM, the module submits the prompt instruction to generate the client utterance of the next turn.  The prompt instruction includes a description of the next state and selected action, along with selected information if the action is type-2.

\subsection{Data Annotation}
\label{sec:data annotation}

\paragraph{Annotation of AnnoMI Dataset.} Our framework assumes that state transition diagram and client profile knowledge exist to develop the state transition, action selection, information selection, and response generation modules. In the absence of available expert-curated knowledge and to allow the framework to adapt to different counseling approaches, we choose to derive these knowledge from AnnoMI~\citep{wu2022anno} for the purpose of implementing and evaluating the framework. We finally select 86 clients and their sessions from the dataset\footnote{The detailed description is provided in Appendix~\ref{app:data annotation}.} then utilized GPT-4\footnote{GPT-4: \texttt{gpt-4-0125}.} to annotate utterances of these selected sessions.
Other than the purpose of framework implementation, we also aim to perform the annotation as accurate as possible so that one can automatically evaluate the consistency of profiles and receptivity of clients in generated sessions against their ground truth ones, and analyze the distribution of client states and actions.

\paragraph{Annotation of Client Profile, States, Actions and Receptivity.} We prompt GPT-4 to summarize the four profile components: persona, motivation, beliefs, and preferred change plan, for the client given a counseling session. These prompts can be found in Table~\ref{tab:profile annotation}.  We also design prompts to annotate the client's state and action (conditioned on the state) at the utterance level (see Tables~\ref{tab:state annotation} and \ref{tab:action annotation} respectively). To determine a client's receptivity in a counseling session, we provide GPT-4 the session labeled with client states and actions followed by prompting GPT-4 to assign a receptivity score (between 1 and 5 as shown in Table~\ref{tab:receptivity annotation}). We repeat five rounds of such assignments and obtain the average receptivity score. We take the floor of the average score to obtain the final receptivity integer score. 

\paragraph{Annotation Validity} To determine the validity of above GPT-4 annotations, two annotators manually reviewed the annotations together and derived the commonly agreed annotation labels.  Against these manually derived ground truth annotations, the GPT-4 annotations achieve high accuracy rates: 87.31\% for states, 85.20\% for actions, and 80.32\% for receptivity scores. The precision and recall of persona annotation are 81.93\% and 80.07\% respectively, that of belief annotation are 77.48\% and 77.38\% respectively. The precision and recall of motivation annotation are 82.72\% and 79.98\% respectively, while that of plan annotation are 78.38\% and 76.24\% respectively. Furthermore, all profile items are factually accurate. Overall, these results confirm the high reliability of GPT-4 annotation method which makes it feasible to automate the evaluation of generated sessions. 

\paragraph{Client Behavior Analysis in AnnoMI sessions.} We analyze the our annotations to reveal client behaviors in AnnoMI sessions. Most clients, ~70\% of them, demonstrate moderate receptivity (score=2 to 4), while a few exhibit high (score=5) or low receptivity (score=1). Clients with higher receptivity tend to use neutral or change talk more often. Conversely, clients with lower receptivity are harder to move to the Contemplation state, requiring more effort from the counselor. Some actions, such as Deny, Downplay, and Reject, are negatively associated with receptivity. The details are provided in Figures~\ref{fig:suggest_annotation} and \ref{fig:action_distribution}.

%% file: sections/experiment.tex
\section{Experiments}
\label{sec:experiments}

Our experiment focuses on evaluating the ability to demonstrate different aspects of consistency for clients simulated using our framework and other methods. First, we introduce the experiment setup and baselines. Next, we report the methods' performance including profile consistency and turn-level response quality. Furthermore, we analyze the similarity between simulated and real sessions in terms of session length, receptivity distribution, motivation rate, and action distribution. Finally, we conduct expert evaluation to highlight the realism and performance of our framework. All the prompts and their description can be found under the supplementary materials.

\subsection{Experiment Setup}
\label{sec:setting}

\paragraph{Counselor and moderator agents.}
Similar to \citet{tu2024towards}, our experiment deploys two other  agents to generate counseling sessions with the simulated clients to be evaluated\footnote{We employ \texttt{gpt-3.5-turbo-0125} as the LLM backbone.}, namely: the {\em  counselor} and {\em moderator} agents. 
While the client agents of different profiles can be simulated using different methods, only one counselor agent and one moderator agent are used. 
The counselor agent assumes the role of an empathetic advisor, trying to identify the client's behavior problem and motivations, thereby facilitating behavior change and making change plans.  Inspired by previous works~\citep{chiu2024computational,wang2024towards,yosef2024assessing}, this agent is developed by prompting the LLM with the basic knowledge about MI and the different stages of counseling (See Table~\ref{tab:couneslor prompt}).  Each counseling session begins with the counselor utterance: ``\textit{Hello. How are you?}'' The client then responds with ``\textit{I am good. What about you?}''. The counselor continues to generate the next utterance to continue the session. The process continues until the moderator decides to end the session.  The above counselor simulation has been validated to work well in terms of therapeutic alliance and skill employment.  

The moderator oversees the dialogue to decide when the session should end.  This decision is made when any of the three conditions is met, namely: 1) the counselor stops persuading the client; 2) the client agrees to a plan of action; or 3) the session reaches its maximum length of 100 turns, similar to the maximum number of turns in AnnoMI sessions.  We implement the moderator agent by prompting LLM with the above end conditions (see Table~\ref{tab:moderator prompt}). 

\paragraph{Client simulation methods.} We evaluate our framework-based client simulation method against four state-of-the-art LLM-based client simulation methods, namely: (a) \textbf{Base Method} which include only the behavioral problem of client in the LLM prompt to generate the next utterance.  This method relies on behavior of the underlying LLMs and has been employed in emotion support conversation task~\citep{deng2023plug, deng2023prompting} (see Table~\ref{tab:baseclient prompt}); (b) \textbf{Example-based Method} which provides a real counseling session exemplar involving the same client and prompts the LLM to simulate the same client talking to a different counselor in a parallel universe~\citep{chiu2024computational} (see Table~\ref{tab:example based prompt}); (c) \textbf{Profile-based Method} which prompts the LLM to simulate the client based on the given client profile~\citep{yosef2024assessing,wang2024patient} (see Table~\ref{tab:profile based prompt}); and (d) \textbf{Pro+Act-based Method} which provides both the client profile and the description of all actions in a LLM prompt to generate the next utterance~\citep{zhang2024strength} (see Table~\ref{tab:action based prompt}). This method however does not involve state-tracking and state-dependent utterance generation. 
 
To obtain more reliable results, each method generates three sessions for each client profile of AnnoMI. We thus have 258 (= 86 $\times$ 3) generated sessions for each method. We then average the profile-specific metric results for the three sessions/client before deriving the average across all the client profiles. 

\subsection{Automated Evaluation of Profile and Receptivity Consistency}
\label{sec:profile consistency}

\begin{table}[tb]
\centering
\resizebox{0.47\textwidth}{!}{
\begin{tabular}{lrrrrc}
\toprule
              & PE$\uparrow$    & MO$\uparrow$         & BE$\uparrow$      & CP$\uparrow$       & RE$\uparrow$  \\ \midrule
Base          &  9.01       & 16.17             & 12.15         &  9.30       & $-0.31^{**}$  \\
Example-based & 53.68       & 45.73             & 45.55         & 33.53       & $0.25^*$    \\
Profile-based & 61.97       & 53.44             & 67.17         & 54.67       & $0.31^{*}$   \\
Pro+Act-based & 67.09       & 55.33             & 68.60         & 57.17       & $0.33^{**}$   \\ \hline
Ours          & \textbf{70.57}   & \textbf{73.37}   & \textbf{71.70}  & \textbf{68.51} & \textbf{0.58}$^{**}$   \\ \bottomrule
\end{tabular}}
\caption{Consistency of Personas (PE), Motivation (MO), Beliefs (BE), and Preferred Change Plans (CP), and Receptivity (RE). High consistency indicates that the client profile information in the generated sessions match the original profile information well. Persona, Motivation, Belief and Plan consistencies are measured by {\em entailment}. Receptivity consistency is measured by {\em Spearman's Correlation} between the client receptivity scores of the generated sessions and those of the original sessions. The superscript ``*'' denotes a p-value less than 0.1, while ``**'' indicates a p-value less than 0.05.}
\label{tab:consistency}
\end{table}

Our automated evaluation includes four aspects of profile consistency, i.e., personas, motivation, beliefs, and preferred change plan, of the simulated clients. It is designed to evaluate a large number of clients and their MI counseling sessions without much human expert effort and cost.
Following the same pipeline for AnnoMI data annotation, we obtain the above four profile components from each generated session and assess if each component (e.g., persona) is entailed in the corresponding profile component from the original AnnoMI session. Specifically, we employ GPT-4 to perform entailment assessment in a few-shot manner. Additionally, for profile components not found in the original session (usually the change plan component), negative entailment outcome is always assigned 
We then define the consistency score of the client simulation method for a profile component by the proportion of sessions with positive entailment for the component. As shown in Table~\ref{tab:consistency}, our method outperforms the rest across all profile components.  This can be attributed to its stronger state control and information selection mechanism. With state control, the motivation and change plan will be compared with the corresponding client profile components. Therefore, they should be consistent when the client changes state. On the other hand, based on the information selection mechanism, only the information in the given profile will be expressed. Thus, the exposed personas and beliefs should be from the profile.  Base method performs the worst, generating not more than 16.17\% of the sessions consistent with the ground truth client profiles across the four aspects. This is reasonable, as no reference profiles were provided. Example-based method also does not perform well, possibly due to the LLM failing to consistently simulate specific personas, beliefs, motivations change plans in lengthy sessions compared to profiles.  Profile-based and Pro+Act-based methods perform rather well in persona but less so in other profile components. These two methods may accept false motivations or plans non-existent in the client profiles. 

For receptivity, we compute the Spearman's correlation between the receptivity scores of simulated clients and clients in original sessions. We follow the client profile construction step to derive the receptivity scores of the simulated clients. As shown in Table~\ref{tab:consistency}, our method outperforms the other baselines with a moderate correlation score. The Base method returns a negative correlation due to not incorporating receptivity. None of Example-based, Profile-based and Pro+Act-based methods can simulate the client with the appropriate receptivity. As shown in Table~\ref{tab:population consistency}, Our client simulation method demonstrates average receptivity (Avg Rec), motivation rate (MR@20) and turns to motivate (Avg MS) similar to that of real sessions.  Clients simulated by other methods in contrast show overly receptive behavior towards the counselor agent as shown by their significantly higher average receptivity, higher motivation rate and fewer turns to motivate.  Only small deviation of receptivity is also observed among them. Among the baselines, Pro+Act-based method can simulate more diverse receptivity by considering action-related instructions.  

\begin{table}[tb]
\centering
\resizebox{0.47\textwidth}{!}{
\begin{tabular}{lrrrr}
\toprule
               & Avg Rec            & MR@20             & Avg MS       & Act KL $\downarrow$        \\ \midrule
Base           & 4.42$_{\pm0.47}$    & 1.00              & 6.60          & 0.39              \\
Example-based  & 4.08$_{\pm0.63}$    & 1.00              & 7.60          & 0.24              \\
Profile-based  & 4.12$_{\pm0.64}$    & 0.96              & 9.76          & 0.15             \\
Pro+Act-based  & 3.86$_{\pm1.01}$    & 0.94              & 9.93          & 0.13              \\ \hline
Ours           & 3.32$_{\pm1.15}$    & 0.69              & 18.60         & 0.06             \\ \hline
Real           & 3.27$_{\pm1.12}$    & 0.48              & 27.56         & 0.00              \\ \bottomrule
\end{tabular}}
\caption{Average receptivity level (Avg Rec), motivation rate at the 20th turns(MR@20), average motivation step (Avg MS), and action KL divergence (Act KL). Our method shows lower receptivity, lower motivation rate (or more motivation steps), and higher action diversity than other methods, which are more aligned with the real sessions.}
\label{tab:population consistency}
\end{table}

\subsection{Session Length and Action Distribution Analysis}
\label{sec:population consistency}

\begin{figure*}[tb]
    \centering
    \includegraphics[width=\textwidth]{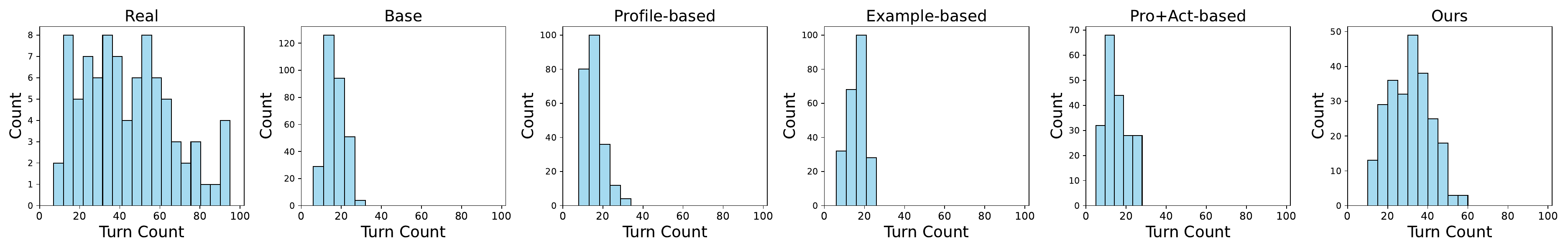}
    \caption{Distribution of turn count (Count) for various clients. The length of AnnoMI counseling sessions is diverse and generally longer, while simulated sessions tend to have fewer than 50 turns.} 
    \label{fig:length distribution}
\end{figure*}


\paragraph{Length of Counseling Sessions.} As shown in Figure~\ref{fig:length distribution}, the length of real counseling sessions ranges from 10 to 100 turns. Most generated ones contain however fewer than 50 turns. Nevertheless, the sessions generated by our method exhibit slightly longer length than previous works. We found that previous methods tend to incorporate more profile information in one utterance as they lack the information selection module. They also prone to having clients motivated to change without much counselor's effort or accept motivations and plans not aligned to their given profiles. In contrast, the shorter sessions generated by our method are mainly due to the counselors giving up when they fail to motivate the simulated clients in the Precontemplation state. This can be explained by our method limiting state changes and leakage of profile information to the counselor through state transition and information selection controls. Without the counselor putting effort into knowing the clients and practising effective MI approach, the counseling session is expected to make little progress. This also highlights the importance of consistent simulated clients in training counselors.


\paragraph{Action Distributions.} We finally evaluate the KL divergence of action distributions between the simulated clients and real clients for the different methods. As shown in Table~\ref{tab:population consistency}, our method demonstrates very small KL divergence which implies that the simulated clients adopt an overall distribution of actions very similar to that of real clients. Other baseline methods, on the other hand, see higher KL divergences as they do not incorporate the real distribution of client actions to guide action selection in realistic counseling scenarios nor employ an action sampling strategy to mitigate selection bias.

\subsection{Expert Evaluation}
\label{sec:exp_eval}

In this section, we further conduct expert evaluation on a small set of sessions generated by a few best performing client simulation methods. Other than profile and receptivity consistency evaluation, we also introduce expert judgment of whether the simulated clients are similar to human clients.  

\paragraph{Profile Consistency} In addition to the GPT-based automated evaluation, we assess the consistency and realism of the simulated clients by experts. We employ four of our co-authors, who are experts in psychology and experienced in MI counseling, as annotators. Their background enables them to comprehend and annotate typical client behaviors. We randomly select six clients and corresponding real sessions from AnnoMI, then employ Profile-Based, Pro+Act-Based, and our methods to simulate the client’s interaction with the counselor agent to generate sessions.\footnote{Given the performance from GPT-based evaluation (Table~\ref{tab:consistency}), we only adopt the Profile-based and Pro+Act-based methods for comparison. We disregard the Base and Example-based methods due to their inferior performance in GPT-based evaluation} For each generated session, we assign three experts to annotate the consistency of profile components, including persona, beliefs, motivation, and plan, compared to the given profile. Each score ranges from 1 (low consistency) to 5 (high consistency).\footnote{More details about expert evaluation can be found in Appendix~\ref{app:human evaluation}.} Finally, we collect a total of 288 annotation scores, derived from 6 clients $\times$ (sessions generated using 3 methods + 1 observed session) $\times$ 3 annotators $\times$ 4 components.  
As shown in Table~\ref{tab:expert_evaluation_consistency}, the results demonstrate that our method outperforms the baselines, which aligns with the GPT-based evaluation. However, there is still a gap between the simulated client and the real client, indicating the potential for future improvement.

\begin{table}[tb]
\centering
\resizebox{0.48\textwidth}{!}{
\begin{tabular}{cccccc}
\toprule
\multirow{2}{*}{} & \multicolumn{4}{c}{Consistency$\uparrow$}             & \multirow{2}{*}{Realism$\uparrow$} \\ \cline{2-5}
                  & Personas & Beliefs & Motivation  & Plan       &                          \\ \midrule
Profile-Based     & 2.61     & 2.00    & 2.61       & 1.56      & 2.38                     \\
Pro+Act-Based     & 2.65     & 2.22    & 2.78       & 1.56      & 2.50                     \\
Our               & \textbf{3.33}     & \textbf{2.89}    & \textbf{3.00}       & \textbf{2.27}  & \textbf{3.16}          \\ \hline
Real              & 4.72    & 4.67   & 4.56       & 4.61      & 4.72                     \\ \bottomrule
\end{tabular}}
\caption{Expert Evaluation on Profile Consistency and Realism for Client Simulation. Our method demonstrates superior performance compared to the two baselines, with a p-value less than 0.05. Nevertheless, the real client consistently outperforms all simulation methods with a p-value less than 0.01. The overall Kappa score of experts evaluation is 0.61 (p-value $\le 0.05$) indicating moderate agreement.}
\label{tab:expert_evaluation_consistency}
\end{table}

\paragraph{Realism} Furthermore, we instruct the experts to assess the realism of client behavior. The experts are tasked with comparing the client’s behavior during the given session with the behavior of actual human client in the real world. We employ the same sessions in consistency evaluation for realism evaluation, which consists of 6 clients and 4 sessions per client. We randomly assign three experts to annotate the realism of client for each session, resulting in a total of 72 annotation scores. The annotation schema employs a five-point Likert scale, ranging from 1 (low realism) to 5 (high realism). The final results are presented in Table~\ref{tab:expert_evaluation_receptivity}. Our simulated clients exhibit significantly superior performance compared to other baselines. Based on the annotation, the primary disadvantage of other baselines is that clients exhibit excessively compromised behaviors, which are unrealistic in the real world.

\paragraph{Receptivity Consistency} We further substantiate our claim that our method can adjust behaviors in accordance with receptivity. To achieve this, we have developed a framework for expert evaluation of receptivity consistency. We randomly select four clients (and their profile components) and generate for each client three counseling sessions covering low receptivity score (1), middle receptivity score (3), and high receptivity score (5). We eliminate the Profile-based method, as the GPT-based evaluations (Table~\ref{tab:population consistency}) demonstrate its performance with high receptivity regardless of the given receptivity in the profile. On the other hand, since the GPT-based evaluation (Table~\ref{tab:population consistency}) suggests that the Pro+Act-based methods exhibit lower and more diverse receptivity compared to other baselines, we incorporate it for comparison with our method. For each session, we randomly assign two experts to annotate the receptivity level of the client in the given session. To facilitate the expert’s annotation and enhance agreement between expert and annotated scores, we simplify the scoring system from five points to three points, categorizing it as low (1), middle (3), and high (5). Finally, we obtain a total of 48 annotation scores (4 clients $\times$ 3 receptivity levels $\times$ 2 methods $\times$ two annotators). Subsequently, similar to the evaluation presented in Tables~\ref{tab:consistency} and \ref{tab:population consistency}, we present the average annotated receptivity score and the Spearman correlation between annotated receptivity scores and the ground truth receptivity scores.

As shown in Table~\ref{tab:expert_evaluation_receptivity}, the sessions generated by our framework achieve the receptivity closer to the assigned one. Almost all sessions generated by the Pro+Act method are annotated as high receptivity, suggesting that the baselines are excessively compliant, disregarding the given receptivity. The receptivity of sessions generated by our framework is highly correlated with the assigned receptivity, indicating the effectiveness of our receptivity control mechanism. However, the average receptivity of our highly receptive client is not sufficiently high, which may be attributed to the imposition of a stringent motivation condition. Only when the counselor touches the inherent motivation of the client will the client be motivated. Consequently, the client may consistently remain in the Precontemplation state, particularly when the counselor lacks the requisite skills.

\begin{table}[tb]
\centering
\footnotesize
\begin{tabular}{ccccc}
\toprule
           & \multicolumn{3}{c}{Avg. Receptivity} & {Receptivity$\uparrow$} \\ 
           \cline{2-4}
           & 1.0          & 3.0          & 5.0         & Correlation                              \\\midrule
Pro+Act Based & 5.0          & 4.3        & 5.0          & 0.00                    \\
Ours       & 1.3        & 3.0          & 4.3         & 0.86                \\ \bottomrule
\end{tabular}
\caption{Expert Evaluation of Receptivity Consistency. Experts assesses the receptivity across various receptivity levels, with the average receptivity for each level presented. Additionally, Spearman Correspondence is calculated to quantify the correlation between the expert’s ratings and the assigned receptivity levels. The Kappa coefficient between expert annotations is 0.87 (p-value $\le 0.001$), indicating high agreement. Specifically, the p-value of Spearman Correspondence is 1.0 for the Pro+Act-Based method, indicating no relationship in the population, while 0.0003 for our method, indicating high significance for the relationship.}
\label{tab:expert_evaluation_receptivity}
\end{table}

%% file: sections/data_annotation.tex
\section{Data Annotation and Analysis}
\label{app:data annotation}

The analysis of both observed (i.e., AnnoMI) and generated MI counseling sessions requires them to be annotated at the session level so as to obtain the client's profile components (i.e., personas, beliefs, motivation, and acceptable change plan) and receptivity.  Tables~\ref{tab:profile annotation} and \ref{tab:receptivity annotation} show corresponding prompts.  We also develop prompts to annotate at a specific point of a session to obtain the client's state and action at that point (see Table~\ref{tab:state annotation} and \ref{tab:action annotation}).  Finally, we create the prompts to determine entailment between generated and ground truth profile components (see Table~\ref{tab:entailment annotation}).  
GPT-4, one of the top proprietary LLMs, is chosen to be the target LLM to perform annotation of the AnnoMI and generated sessions.

In the following, we report the analysis results after annotating the 86 selected AnnoMI sessions.

\paragraph{AnnoMI Dataset} We utilize the AnnoMI dataset~\citep{wu2022anno}, which comprises 133 conversations spanning a diverse range of behavior topics, including ``reducing alcohol consumption'' and ``smoking cessation.'' Each conversation was transcribed from an expert-generated demonstration video and subsequently labeled as demonstrating high- or low-quality MI based on the video title and description. The 110 dialogues illustrating high-quality MI, comprising over 8,800 utterances in total, are considered as candidates. Consequently, we eliminate the sessions that are incomplete and those involving clients with insufficient profile information. To be noticed, the AnnoMI dataset is compiled from YouTube videos depicting MI counseling sessions between actors. However, it remains widely utilized due to restricted access to real counseling sessions. Given the absence of ideal datasets, our framework offers a solution to the scarcity of data by simulating clients interacting with counselors, thus bypassing ethical and privacy issues. Additionally, the primary use of the AnnoMI dataset is to extract client profiles and corresponding counseling sessions for evaluation purposes. Although these sessions are not “real”, they are still consistent with the client profiles and serve as valuable evaluation dataset.

\begin{figure*}[t]
    \centering
    \includegraphics[width=\textwidth]{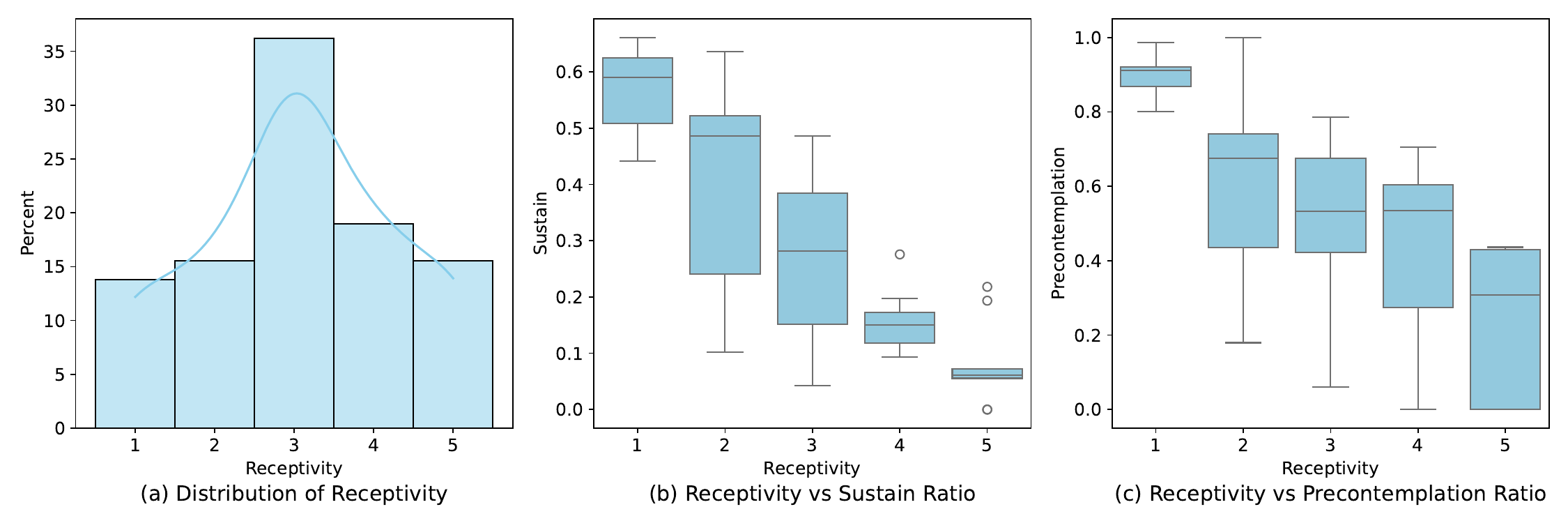}
    \caption{The distribution of receptivity and the relation between receptivity and sustain ratio and precontemplation.}
    \label{fig:suggest_annotation}
\end{figure*}

\begin{figure*}[tb]
    \centering
    \includegraphics[width=\textwidth]{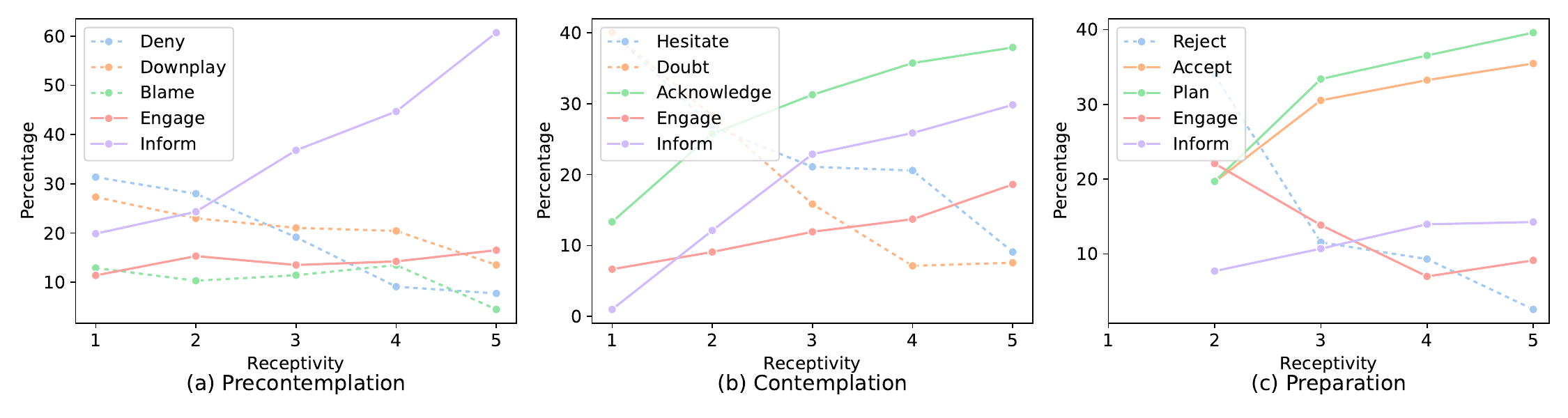}
    \caption{Proportions of actions for different receptivity scores in different states. Actions negatively associated with receptivity are represented in dashed lines.}
    \label{fig:action_distribution}
\end{figure*}

\paragraph{Analysis of receptivity, associated talk types, and client states.} As shown in Figure~\ref{fig:suggest_annotation}(a), most clients demonstrate moderate receptivity. Few clients have high and low receptivities.  We next analyse the client utterances labeled with talk-types ``change talk'', ``neutral'' and ``sustain talk''. utterances annotated with ``Change talk'' express an interest to change behavior, while those annotated with ``sustain talk'' are the opposite~\citep{hoang2024can, miller2012motivational}. Figure~\ref{fig:suggest_annotation}(b) shows that the ratio of sustain talk-labeled utterances has a negative relationship with receptivity, i.e., clients with higher receptivity tend to use neutral or change talk more often.  Figure~\ref{fig:suggest_annotation}(c) shows that the proportion of client utterances in Precontemplation state also has a negative relationship with receptivity as clients in this state are reluctant and/or the counselor is not able to effect behavior change. Clients with lower receptivity are harder to reach the contemplation state, thus requiring more effort from the counselor.

\paragraph{Analysis of receptivity and associated actions.} Figure~\ref{fig:action_distribution}(a) illustrates the proportion of actions in different client states across different receptivity scores. Considering the utterances of clients who are in the precontemplation state, the proportion of utterances annotated with Inform action increases with receptivity due to the more receptive clients providing information about themselves.  The proportion with Engage action also increases slightly.  On the other hand, the proportions of Deny, Downplay and Blame actions reduces as receptivity increases. Similar findings also apply to Figures~\ref{fig:action_distribution}(b) and (c).  For easy reading, the actions negatively associated with receptivity are shown in dashed lines.  
Finally, Inform and Engage are two actions found in all the three states. Inform is most predominant in the Precontemplation state implying that clients are more likely to share profile information during Precontemplation and adopt more varied actions after they transit to Contemplation and Preparation states. 

\begin{table*}[tb]
\begin{tabularx}{\textwidth}{X}
\toprule
\#\# Task \newline Your task is to identify the client's profile based on the provided counseling conversation. Focus on the following aspects: \newline  - **Persona**: Include personal details such as recent events, family relationships, and occupation. Summarize each aspect in concise sentences. \newline - **Behavioral Problem**: Identify a key problematic behavior (e.g., drinking, smoking). Describe it succinctly in a single phrase, focusing on the primary issue. \newline - **Motivation (Optional): Explain why the client wants to change this behavior (due to family, health, work, etc.) in one sentence, focusing on the main reason. Leave this section blank if the client shows no motivation to change.  \newline - **Beliefs**: Detail the client's inner beliefs toward change, such as downplaying the behavior's frequency, blaming others, or doubting the benefits of change. Provide a sentence for each reason. \newline  - **Acceptable Plan** (Option): Describe any plans the client is willing to adopt or consider, like reducing frequency, altering the environment, or seeking help. Use one sentence for each plan. Leave this blank if the client accepts no plans. \newline  \#\# Output Format  \newline Return the client profile in JSON format as illustrated below:  \newline [example]  \newline \#\# Given Conversation [conversation]  \newline \#\# Instruction  \newline Based on the conversation, provide the client's profile in JSON format.
\\ \bottomrule
\end{tabularx}
\caption{Prompt for GPT-4 to annotate the profile of client in given conversation or session ([conversation] is to be replaced by the session content).}
\label{tab:profile annotation}
\end{table*}

\begin{table*}[tb]
\begin{tabularx}{\textwidth}{X}
\toprule
\#\# Task
Your task is to read through the provided counseling conversation and determine which stage of change the client is in based on the Transtheoretical Model of behavior change. The stages you will focus on are Precontemplation, Contemplation, and Preparation. \newline \#\# Definitions of Stages \newline Precontemplation: The client is not yet acknowledging that there is a problem behavior that needs to be changed. They may be defensive, dismissive, or indifferent about the topic of change. \newline Contemplation: The client acknowledges that there is a problem and begins to think about the possibility of change. However, they are not yet committed to taking action but are more open to discussing the pros and cons of their behavior. \newline Preparation: The client is planning to change and is likely to start taking steps soon. They might begin to set goals, seek information, or plan out the changes they intend to make.  \newline \#\# Instructions \newline Read the Conversation: Carefully read the entire conversation to understand the context and content of the discussion.
\newline Identify Statements and Attitudes: Pay close attention to the client’s statements and attitudes towards change. Look for keywords or phrases that indicate their stage of change according to the definitions provided. \newline \#\# Determine the Stage \newline - If the client shows no recognition of the problem or need for change, categorize them as Precontemplation.  \newline - If the client acknowledges the problem and discusses thoughts about possibly changing, without commitment, categorize them as Contemplation. \newline - If the client talks about specific plans or the intention to change soon, categorize them as Preparation. \newline \#\# Justify Your Choice  \newline Provide a brief justification for the stage you have assigned based on specific parts of the [conversation]. Include direct quotes or clear references to the conversation to support your decision.  \newline \#\# Given Conversation: [conversation]  \newline Annotate the state of each utterance of client.
\\ \bottomrule
\end{tabularx}
\caption{Prompt for GPT-4 to annotate the latest stage (or state) of client in given conversation ([conversation] is to be replaced by the session which as be complete or partially complete).}
\label{tab:state annotation}
\end{table*}

\begin{table*}[tb]
\begin{tabularx}{\textwidth}{X}
\toprule
\#\# Task \newline Your task is to annotate the action of the client for each provided conversation snippet. For each snippet, focus on the client's last utterance and choose the most appropriate action from the given set of action. \newline  \#\# Action Options \newline [options] \newline \#\# Instructions \newline - Read the Conversation Snippet: Carefully read through the provided snippet. \newline - Focus on the Last Utterance: Concentrate on the client's last utterance. \newline - Choose the Action: Select the action that best describes the client's last utterance from the given set of actions. \newline  \#\# Response Format \newline Analysis of Client's Action: Provide a brief analysis of the client's last utterance and the chosen action. \newline Chosen Action: [Selected action] \newline For example: \newline Analysis of Client's Action: The client shows uncertainty about the effectiveness of potential stress management strategies, indicating hesitation.  \newline Chosen Action: Hesitate  \newline Provided Conversation Snippet for Annotation [conversation]  \newline What is the most appropriate action that describes the client's last utterance ([last utterance]) in the conversation snippet? Provide a brief analysis to support your choice.
\\ \bottomrule
\end{tabularx}
\caption{Prompt for GPT-4 to assign  from a list of possible actions [options] the action of the last client's utterance ([last utterance]) in the given conversation snippet ([conversation]).}
\label{tab:action annotation}
\end{table*}

\begin{table*}[tb]
\begin{tabularx}{\textwidth}{X}
\toprule
\#\# Task \newline Your task is to assess the client's receptivity in a given counseling conversation by focusing on the client's behaviors, openness, and responsiveness to suggestions. Determine how easily the counselor can motivate the client to change. \newline \#\# Guidelines \newline The client with higher receptivity tends to openly share information, accept suggestions, and show high confidence in their ability to change. The client with lower receptivity tends to resist suggestions, deny the need for change, downplay the importance of change, and exhibit other passive attitudes. \newline \#\# Scoring Receptivity \newline - **High Receptivity (5)**: A client with high receptivity consistently demonstrates behaviors such as frequently and openly sharing personal thoughts and feelings, actively engaging in discussions about change, quickly and positively responding to suggestions, and displaying strong self-belief in their ability to make changes. \newline - **Moderately High Receptivity (4)**: Clients with moderately high receptivity are generally open and forthcoming, mostly receptive to change discussions, respond well to suggestions with occasional need for reinforcement, and demonstrate good self-belief with some need for reassurance. \newline - **Moderate Receptivity (3)**: Clients with moderate receptivity show a balance between sharing and withholding information, exhibit mixed interest in change discussions, respond to suggestions neutrally, and display a balanced view of their ability to change. \newline - **Moderately Low Receptivity (2)**: Clients with moderately low receptivity are characterized by reluctance to share personal information, limited interest in change discussions, resistance to suggestions, and low self-belief with frequent expression of doubts. \newline - **Low Receptivity (1)**: Clients with low receptivity consistently withhold information, avoid discussions about change, strongly resist suggestions, and show very low self-belief, often highlighting reasons why change is not possible. \newline \#\# Important Note \newline Remember that the client may be motivated to change by the end of the conversation due to the counselor's efforts. However, your assessment should be based on the client's behaviors throughout the conversation especially before issue acknowledgement. \newline \#\# Response Format \newline Analysis of Client's Behavior: Provide a detailed analysis of the client's behaviors during the conversation. \newline Receptivity Score: Assign a final score based on the observed behaviors using the provided scoring system. \newline For Example:  \newline Analysis of Client's Behavior: The client initially exhibited resistance to change by denying the need for it. As the conversation progressed, the client showed some openness by considering small changes but remained hesitant and doubtful. Overall, the client demonstrated a low receptivity behaviors. \newline Receptivity Score: 1 (Low receptivity) \newline \#\# Provided Conversation for Evaluation \newline [conversation]  \newline What is the client's receptivity score based on the observed behaviors in the conversation? Provide a detailed analysis of the client's behaviors to support your assessment.
\\ \bottomrule
\end{tabularx}
\caption{Prompt for GPT-4 to annotate the receptivity score of client (1 to 5) in given conversation snippet ([conversation] will be replaced by the counseling session).}
\label{tab:receptivity annotation}
\end{table*}

\begin{table*}[tb]
\begin{tabularx}{\textwidth}{X}
\toprule
\#\# Task \newline You task is to identify whether the premise entails the hypothesis. The answer should be exact 'entail' or 'not entail'. \newline \#\# Premise [profile] \newline \#\# Hypothesis [component]
\\ \bottomrule
\end{tabularx}
\caption{Prompt for the GPT-4 to evaluate the entailment between each generated profile component ([component]) and given profile ([profile]).}
\label{tab:entailment annotation}
\end{table*}

%% file: sections/human_annotation
\section{Expert Annotation}
\label{app:human evaluation}

\subsection{Demographics for Annotators}

Our annotation team comprises four experts who are co-authors of this paper. Three of them hold postgraduate degrees in counseling and one is a senior-year university student pursuing a psychology major. They all have relevant experience in mental health counseling.

\subsection{Annotation Instruction}

Table~\ref{tab:human_instruction1} presents the profile consistency rating score descriptions and client profile components for an expert to evaluate the profile consistency scores of a generated counseling session in all the four profile aspects (i.e., persona, beliefs, motivation, and acceptable plan). Table~\ref{tab:human_instruction2} shows the realism rating score descriptions for an expert to assess the realism of client in a generated session. Finally, Table~\ref{tab:human_instruction3} shows the score descriptions for an expert to determine the receptivity level of client in a generated counseling session.

\begin{table*}[tb]
\begin{tabularx}{\textwidth}{X}
\toprule
\textbf{Consistency Rating (To rate how consistent is the client with the four aspects of profile)}\newline
5 (High): Perfectly aligns with profile across all aspects \newline
4 (Moderately High): Mostly aligns with profile with minor variations \newline
3 (Moderate): Partially aligns with profile with some discrepancies \newline
2 (Moderately Low): Limited alignment with profile with multiple discrepancies \newline
1 (Low): Consistently contradicts profile information \newline
\textbf{CLIENT PROFILE} \newline
\textbf{Behavior:} drinking alcohol \newline
\textbf{Personas:}\newline
- The client sneaked out with a friend to the park to has a couple of beers.\newline
- The client managed to avoid getting into trouble as he threw the beers away before the policy saw them.\newline
- If the the client's mother found out he was in trouble because of his drinking, she would kill him.\newline
- The client has an interest in soccer and has set a goal to play in college.\newline
- The client has aspirations of receiving scholarships to support his soccer career.\newline
\textbf{Beliefs:} \newline
- The client believes he can handle the dangerous situation to avoid getting into trouble. \newline
- The client has only drunk once or twice. \newline
- A couple of the client's friends drink too. \newline
\textbf{Motivation:} The client is motivated because of the risk of drinking alcohol in sports, as alcohol would affect the ability to play soccer. \newline
\textbf{Acceptable Plans:} None \newline
\textbf{SESSION} \newline
[conversation]
\\ \bottomrule
\end{tabularx}
\caption{Description of profile consistency rating scores, ground truth client profile, and counseling session for expert annotation.  The expert is required to give a rating for each profile aspect. ([conversation] will be replaced by the real or generated session.)}
\label{tab:human_instruction1}
\end{table*}

\begin{table*}[tb]
\begin{tabularx}{\textwidth}{X}
\toprule
\textbf{Realism Rating (To rate how realistic is the client compared to a human client)} \newline
5 (High): Shows authentic responses with natural flow and rich personal context\newline
4 (Moderately High): Shows mostly genuine responses with occasional scripted moments\newline
3 (Moderate): Shows mixed authentic and artificial responses\newline
2 (Moderately Low): Shows mostly artificial responses with few authentic moments\newline
1 (Low): Shows completely artificial responses with no authentic elements\newline
\textbf{SESSION}\newline
[conversation]
\\ \bottomrule
\end{tabularx}
\caption{Description of realism rating scores, and counseling session for expert annotation. ([conversation] will be replaced by the real or generated session.)}
\label{tab:human_instruction2}
\end{table*}

\begin{table*}[tb]
\begin{tabularx}{\textwidth}{X}
\toprule
\textbf{Receptivity Rating (To rate how receptive is the client in the session)} \newline
5 (High): Actively engages, shares openly, and responds positively \newline
3 (Moderate): Shows balanced sharing and mixed response patterns \newline
1 (Low): Consistently withholds and strongly resists engagement \newline
\textbf{SESSION} \newline
[conversation]
\\ \bottomrule
\end{tabularx}
\caption{Description of receptivity ratings, and counseling session for expert annotation. ([conversation] will be replaced by the real or generated session.)}
\label{tab:human_instruction3}
\end{table*}

%% file: sections/more_evaluation.tex
\section{More Evaluation Results}
\label{app:more evaluation}

\subsection{Automated Turn-Level Evaluation}

\begin{table*}[tb]
\centering
\begin{tabular}{lrrrrrrrr}
\toprule
             & ROUGE-1 & ROUGE-2 & ROUGE-L & BERTScore \\ \midrule
Base         & 13.27   & 2.55    & 10.03   & 85.59     \\
\rowcolor[HTML]{D8D6D6} Example-based & 59.12   & 48.84   & 58.31   & 93.54     \\
Profile-based & 13.79   & 3.18    & 9.90    & 85.50     \\
Pro+Act-based & 17.04   & 4.22    & 13.93   & 86.72     \\ \hline
Ours         & 17.73   & 9.13    & 15.13   & 87.13     \\
Ours*        & \textbf{29.69}   & \textbf{20.19}   & \textbf{28.75}   & \textbf{90.46}  \\ \bottomrule
\end{tabular}
\caption{Results of turn-level evaluation. ``Our'' refers to our client simulation method. ``Ours*'' refers to a variant of our method provided with ground truth states, actions, and selected information. Other than Example-based method, our framework outperforms the remaining baseline methods.}
\label{tab:turn-eval}
\end{table*}

As previous works~\citep{wang2024towards,deng2023prompting,kang2024can} often evaluated generated utterances against ground truth utterances in conversational tasks, we also conduct turn-level evaluation which involves using a simulated client to generate the next client utterance given a prompt that includes all the session content up to the previous utterance. To show the upper bound performance of our simulated client method (i.e., ``Ours''), we provide the results of a variant of our method (i.e., ``Ours*'') that generates the next utterance using the ground truth state, action and selected information. As shown in Table~\ref{tab:turn-eval}, our method outperforms most of the baselines. The Example-based method has been shaded because it has an unfair advantage over others due to its ability to search all the client utterances including the ground truth utterance. Our method variant, Ours*, shows that there are still rooms for Our method to improve in ROUGE-1, ROUGE-2 and ROUGE-L. Our and Our* methods share very similar BERTScore performance. This indicates that the Our method can reproduce responses semantically similar to the ground truth utterances. 

Nevertheless, it is important to note that turn-level evaluation only measures quality of the next utterance assuming that all earlier utterances are known.  This assumption does not hold in a real world counseling session or application where the client and counselor do not follow some written counseling scripts. Hence, a method performing well in this evaluation method may not always do so at the session level.  For example, the Example-based method appears to perform well in the turn-level evaluation purely due to its ability to find the most similar client utterance, indicating that turn-level evaluation may not accurately measure a client's effectiveness in real world scenarios. Hence, consistency evaluation of client profile components in Section~\ref{sec:profile consistency}, and client receptivity are introduced in Section~\ref{sec:exp_eval}. We believe these results can provide a more comprehensive assessment of client performance.

\subsection{Evaluation on Multiple Backbones}

To demonstrate the generalizability of our framework, we employ various LLMs as the backbone to implement our framework. As presented in Table~\ref{tab:results for multiple models}, our method consistently outperforms the baselines across various backbones. The performance of LLMs in simulation varies due to their distinct instruction-following and conversation capabilities. In summary, the \texttt{gpt-4o}, \texttt{llama-70B}, and \texttt{mistral-large} perform comparably to the \texttt{gpt-3.5}, while the smaller LLMs (\texttt{ministral-8B} and \texttt{llama-8B}) exhibit significantly inferior performance.

\begin{table*}[tb]
\centering
\resizebox{\textwidth}{!}{
\begin{tabular}{lrrrrrrrrr}
\toprule
              & Personas$\uparrow$    & Motivation$\uparrow$         & Beliefs$\uparrow$      & Plans$\uparrow$       &$\rho$ $\uparrow$  & Avg. Rec.            & MR@20             & Avg.MS.      & Act. KL $\downarrow$ \\ \midrule
Real          & - & -    & -  & -  & - & 3.27$_{\pm1.12}$    & 0.48              & 27.56         & 0.00              \\ 
\toprule
\multicolumn{10}{c}{\cellcolor[HTML]{EFEFEF}\texttt{gpt-4o-2024-11-20}}                              \\ \midrule
Base          & 9.17        & 16.19             & 12.26         & 9.38       & -0.28   & 4.30$_{\pm0.31}$    & 1.00              & 7.30          & 0.40 \\
Example-based & 56.29       & 49.96             & 50.66         & 37.21       & 0.27   & 3.98$_{\pm0.68}$    & 1.00              & 9.12          & 0.23\\
Profile-based & 69.78       & 56.97             & 70.59         & 57.31       & 0.32   & 4.05$_{\pm0.72}$    & 0.90              & 10.26          & 0.15 \\
Pro+Act-based & 72.12       & 63.43             & 71.57         & 63.89       & 0.35   & 3.63$_{\pm0.97}$    & 0.85              & 11.80          & 0.14\\ \hline
Ours          & \textbf{76.10}   & \textbf{78.57}   & \textbf{76.13}  & \textbf{74.99} & \textbf{0.66}  & 3.24$_{\pm1.21}$    & 0.61              & 19.78         & \textbf{0.05} \\ \bottomrule
\toprule
\multicolumn{10}{c}{\cellcolor[HTML]{EFEFEF}\texttt{Mistral-Large-Instruct-2411}}                              \\ \midrule
Base          &  9.18       & 16.22             & 12.21         & 9.35        & -0.30  & 4.38$_{\pm0.38}$    & 1.00              & 7.22          & 0.38 \\
Example-based & 54.22       & 47.18             & 47.01         & 34.87       & 0.26    & 4.04$_{\pm0.69}$    & 1.00              & 8.06          & 0.23\\
Profile-based & 66.31       & 57.10             & 69.20         & 56.85       & 0.33   & 4.22$_{\pm0.63}$    & 0.93              & 10.05          & 0.15 \\
Pro+Act-based & 70.11       & 60.88             & 71.09         & 60.03       & 0.36   & 3.77$_{\pm0.91}$    & 0.88              & 10.97          & 0.12\\ \hline
Ours          & \textbf{73.21}   & \textbf{76.30}   & \textbf{73.13}  & \textbf{71.67} & \textbf{0.62}  & 3.17$_{\pm1.27}$    & 0.65              & 19.56         & \textbf{0.05} \\ \bottomrule
\toprule
\multicolumn{10}{c}{\cellcolor[HTML]{EFEFEF}\texttt{LLaMA-3.3-70B-Instruct}}                              \\ \midrule
Base          &  9.15       & 16.20             & 12.17         &  9.35       & -0.31  & 4.33$_{\pm0.43}$    & 1.00              & 6.97          & 0.40 \\
Example-based & 54.73       & 47.56             & 46.26         & 34.78       & 0.27    & 4.07$_{\pm0.62}$    & 1.00              & 8.10          & 0.23\\
Profile-based & 67.54       & 55.04             & 68.51         & 57.95       & 0.33   & 4.11$_{\pm0.67}$    & 0.91              & 10.76          & 0.14 \\
Pro+Act-based & 70.08       & 59.43             & 69.98         & 59.81       & 0.35   & 3.82$_{\pm0.98}$    & 0.89              & 10.83          & 0.13\\ \hline
Ours          & \textbf{73.92}   & \textbf{75.90}   & \textbf{73.10}  & \textbf{71.81} & \textbf{0.60}  & 3.20$_{\pm1.08}$    & 0.63              & 19.02         & \textbf{0.06} \\ \bottomrule
\toprule
\multicolumn{10}{c}{\cellcolor[HTML]{EFEFEF}\texttt{gpt-3.5-turbo-0125}}                              \\ \midrule
Base          &  9.01       & 16.17             & 12.15         &  9.30       & -0.31  & 4.42$_{\pm0.47}$    & 1.00              & 6.60          & 0.39 \\
Example-based & 53.68       & 45.73             & 45.55         & 33.53       & 0.25    & 4.08$_{\pm0.63}$    & 1.00              & 7.60          & 0.24\\
Profile-based & 61.97       & 53.44             & 67.17         & 54.67       & 0.31   & 4.12$_{\pm0.64}$    & 0.96              & 9.76          & 0.15 \\
Pro+Act-based & 67.09       & 55.33             & 68.60         & 57.17       & 0.33   & 3.86$_{\pm1.01}$    & 0.94              & 9.93          & 0.13\\ \hline
Ours          & \textbf{70.57}   & \textbf{73.37}   & \textbf{71.70}  & \textbf{68.51} & \textbf{0.58}  & 3.32$_{\pm1.15}$    & 0.69              & 18.60         & \textbf{0.06} \\ \bottomrule
\toprule
\multicolumn{10}{c}{\cellcolor[HTML]{EFEFEF}\texttt{Ministral-8B-Instruct-2410}}                              \\ \midrule
Base          &  9.10       & 16.17             & 12.09         &  9.27       & -0.34  & 4.54$_{\pm0.42}$    & 1.00              & 6.87          & 0.39 \\
Example-based & 50.54       & 40.88             & 39.45         & 28.51       & 0.13    & 4.49$_{\pm0.61}$   & 1.00              & 7.37          & 0.30\\
Profile-based & 56.77       & 48.23             & 61.81         & 48.50       & 0.21   & 4.40$_{\pm0.60}$    & 0.99              & 9.56          & 0.19 \\
Pro+Act-based & 60.39       & 52.78             & 64.91         & 51.73       & 0.28   & 4.09$_{\pm0.91}$    & 0.96              & 9.53          & 0.17\\ \hline
Ours          & \textbf{65.75}   & \textbf{68.70}   & \textbf{68.98}  & \textbf{61.71} & \textbf{0.51}  & 3.58$_{\pm1.23}$    & 0.72              & 17.67         & \textbf{0.10} \\ \bottomrule
\multicolumn{10}{c}{\cellcolor[HTML]{EFEFEF}\texttt{LLaMA-3.1-8B-Instruct}}                              \\ \midrule
Base          &  9.13       & 15.85             & 12.02         &  9.21       & -0.33  & 4.52$_{\pm0.39}$    & 1.00              & 6.75          & 0.38 \\
Example-based & 50.46       & 40.53             & 39.35         & 28.13       & 0.13    & 4.32$_{\pm0.68}$    & 1.00             & 7.22          & 0.31\\
Profile-based & 55.89       & 47.81             & 61.71         & 48.37       & 0.20   & 4.39$_{\pm0.58}$    & 0.99              & 9.72          & 0.20 \\
Pro+Act-based & 60.05       & 51.90             & 64.16         & 50.96       & 0.27   & 4.21$_{\pm0.95}$    & 0.96              & 9.49          & 0.15\\ \hline
Ours          & \textbf{64.57}   & \textbf{67.90}   & \textbf{67.68}  & \textbf{61.68} & \textbf{0.50}  & 3.52$_{\pm1.17}$    & 0.73              & 16.74         & \textbf{0.11} \\ \bottomrule
\end{tabular}}
\caption{Consistency Assessment of Personas, Motivations, Beliefs, and Preferred Change Plans, Spearman’s Correlation of Receptivity ($\rho$), the average receptivity level (Avg. Rec.), motivation rate (MR) in the initial 20 turns, and average motivation step for baselines and our methods based on various LLMs are presented. In all backbones, our method consistently outperforms all baselines.}
\label{tab:results for multiple models}
\end{table*}

\subsection{Employ Multiple Actions in One Response}

Our current study focuses on single action selection due to the observation that the majority of utterances (89.57\% in the AnnoMI dataset) involve only one action. Only 10\% of utterances involve multiple actions. From the perspective of our framework, incorporating more complex actions can be achieved seamlessly with minor modifications. These complex actions can be included in the action set so that they can be selected for utterance generation. Alternatively, multiple actions can be sampled from the action distribution. To demonstrate the capability of our framework in using multiple actions to generate an utterance, we present an example where multiple potential actions are sampled from the action distribution. To determine the number of actions for generating an utterance, we initially sample a number of actions from the observed data (i.e., 89(single):10(double):1(three)) as candidates. Subsequently, we sample actions from the candidate actions based on the action distribution. Finally, we simply combine all the descriptions of sampled actions as instructions. Table~\ref{tab:multiple actions} demonstrates an example where our framework selects multiple actions probabilistically to control more intricate behaviors in a single response. However, it is important to note that almost all responses can be generated within single actions, thus we believe the single action framework can easily extended to simulate the multiple action scenarios. 

\begin{table*}[htbp]
\centering
\begin{tabularx}{\textwidth}{X}
\toprule
\multicolumn{1}{c}{\textbf{Example of Client Employing Multiple Actions}} \\ \midrule                                                                                             
Counselor: Hello. How are you?                                                                                                                                                                                                                                                                                                \\
Client   : I am good. What about you?                                                                                                                                                                                                                                                                                         \\
Counselor: I'm doing well, thank you! Smoking can significantly impact your finances, from the cost of cigarettes to potential health-related expenses and lost income. I'm curious to hear your thoughts on how smoking might be affecting your financial situation. Please share more about that.                           \\
Client ({\color{red}Deny}, {\color{green}Downplay}): {\color{red} You know, I don't really think about the money too much.} {\color{green} Life is busy, and there are so many other things to focus on, like work and my daughter. I just enjoy my routine and don't see smoking as a big issue in the grand scheme.}                                                                                \\
Counselor: It's great that you have a busy life with work and your daughter! I'm curious, how do you think smoking might impact your productivity or relationships at work? Have you noticed any effects on your career or how you feel in the workplace? Remember, it's all about your choices.                              \\
... \\
Client (Blame): Honestly, almost of my friends smoked, it would be totally normal and we could hang out without feeling bad about it.                   \\
Counselor: While smoking seems normal, it poses significant health risks, including respiratory issues and chronic diseases, which can impact your energy and time with your daughter. Choosing healthier habits can enhance your well-being and ensure you can fully engage in those precious moments together.              \\
Client ({\color{red}Engage}, {\color{green} Blame}): {\color{red} I get what you're saying, but honestly, life is so stressful right now. Between work and everything with my daughter, I just need something to help me unwind.} {\color{green} Plus, everyone around me has their own coping mechanisms, so it feels normal to me.}                                                                 \\
Counselor: It sounds like smoking provides a familiar escape for you amidst the chaos of daily life, and it’s understandable to want that comfort. However, consider how quitting could enhance your physical fitness—boosting your energy and flexibility—ultimately making you more present and engaged with your daughter. \\
Client (Downplay): I see your point, but I really don't think my smoking is affecting my daughter. She's too young to understand, and I don't smoke around her anyway. Everyone has their own way to deal with stress, and mine happens to be smoking. It’s just part of my routine.                                                  \\
Counselor: I’m concerned that your smoking could be diminishing your energy and flexibility, potentially impacting your ability to keep up with your daughter’s activities. Quitting could not only improve your physical fitness but also enhance your enjoyment of those special moments together.                          \\
Client (Deny): I think that's a bit dramatic. My daughter loves it when I take her to the park and play games with her. She doesn't care that I smoke – she just loves spending time with me. And I'm right there with her, enjoying it too. Why would quitting make a difference?                                                \\
Counselor: I understand that relaxation is essential for you. However, viewing your stress and tiredness as signs to prioritize your health could guide you towards healthier coping strategies. Reducing alcohol might enhance your energy and overall well-being, allowing you to embrace life with greater vitality.       \\
Client (Downplay): I guess we all have our ways of coping. Honestly, I think I have it under control.                                                                          \\
Counselor: I understand that smoking feels manageable, but it might limit your energy for activities with your daughter. Quitting could enhance your stamina, improve your health, and help you engage more fully in those cherished moments together, supporting both your well-being and your relationship.                 \\
Client ({\color{red} Downplay}, {\color{green} Inform}): {\color{red}I see your point, but I really don’t think my smoking is a big deal. It’s just part of my routine, and we have fun together anyway.} {\color{green}I don’t want my daughter to see it as something negative, but I also feel it doesn’t affect our time together much.}                                                            \\
...                                                                                                               \\ \bottomrule    
\end{tabularx}
\caption{An example demonstrates the ability of our framework to assign multiple actions for one response. This example highlights the extensibility of our framework to more complex control scenarios involving multiple actions within a single response. Each utterance corresponding to a distinct action is distinguished by a unique color.}
\label{tab:multiple actions}
\end{table*}

\subsection{Complex State Transition}

Although our framework primarily employs a unidirectional flow of state tracking, which may limit its practicality, it is still an initial attempt to introduce a state tracking module in client simulation based on observed state flow from the referenced AnnoMI dataset. However, relapse scenarios (i.e., clients returning to pre-contemplation state from contemplation state) were not observed in the AnnoMI data, preventing the derivation of their sampling distribution. To support more complex state flows, we would need to access large real-world counseling datasets, which are scarce. To demonstrate the extensibility of our framework to incorporate complex state flows, we simply employ a random method, where the client relapses to a previous state with probability (set as 0.3) in each turn. If the counselor fails to resolve the client’s hesitation or address the beliefs effectively during Contemplation, the client may relapse to Precontemplation and deny the awareness of change. In the Preparation state, if the counselor fails to motivate the client to make an acceptable plan effectively, the client may relapse to the Contemplation state. Furthermore, since relapse can occur in each turn, longer sessions may be penalized.

Table~\ref{tab:complex transition1} and Table~\ref{tab:complex transition2} demonstrate an example by employing such a complex state transition flow, where the client relapses to Precontemplation when the counselor fails to address the beliefs well. The experiment not only suggests the extensibility of our framework but also demonstrates the powerful controllability of our framework in state transition. It is important to note that such complex state flows could not be observed in the real AnnoMI dataset. Moreover, we believe that the relapse of client's states in real-world scenarios is more likely to occur across sessions instead of within a single session~\citep{hashemzadeh2019transtheoretical,prochaska1997transtheoretical,miller2012motivational}.

\begin{table*}[htbp]
\centering
\begin{tabularx}{\textwidth}{X}
\toprule
\multicolumn{1}{c}{\textbf{Example of Client Employing Complex State Flow (Part I)}} \\ \midrule   
Counselor: Hello. How are you? \\
Client: I am good. What about you? \\
Counselor: I'm glad to hear you're doing well! It’s important that you're here to reflect on your drinking. How do you think your alcohol use impacts your relationships, like with your family or friends? This could be a good opportunity to explore its effects. \\
Client {\color{green}(Precontemplation)}: Well, my friends and I just enjoy our time together, and it's more about socializing than anything else. My family has always had their own things going on, so it's not like my drinking affects them. Plus, everyone else drinks, so it's pretty normal, right? \\
Counselor: It sounds like you see drinking as a normal part of socializing with friends. Tell me more about what you enjoy most during those times together. Have you noticed any effects on how you feel, both physically and mentally, after a night of drinking? \\
Client {\color{green}(Precontemplation)}: I just feel relaxed and it’s a good way to unwind after a long week. Sometimes I might feel a bit sluggish the next day, but I think it’s pretty normal. I don’t see having a few beers as a problem—it's just part of having a good time. \\
Counselor: I understand drinking feels like a normal part of socializing. However, reducing alcohol can improve your strength and flexibility, making activities more enjoyable. Have you noticed how drinking affects your performance in sports or other physical activities with friends? Healthier choices can enhance both fun and fitness! \\
Client {\color{green}(Precontemplation)}: Actually, I've noticed I get a lot more tired after a few drinks, so I try to limit my activity after a night out. But overall, my performance in sports hasn't really been affected by drinking – I still do fine. It's not a big deal, really. \\
Counselor: It’s great that you feel fine in sports, but have you considered how the tiredness from drinking might impact your overall fitness goals? Reducing alcohol could boost your endurance and energy levels, making your workouts more effective. What are your thoughts on that? \\
{\color{blue} Transfer to Contemplation} \\
Client {\color{red} (Contemplation)}: You make a really good point. I hadn't thought about how the tiredness from drinking could really hold me back from achieving my fitness goals. I want to feel more energetic and perform better in my workouts, and I can see how cutting back on alcohol could help with that. I'm definitely motivated to make some changes! If I can improve my endurance and get more out of my training, it would definitely be worth it. Thanks for highlighting that. \\
Counselor: It sounds like your tiredness after drinking could be a signal that your body is ready for a change. By reducing alcohol, you might not only boost your energy but also enhance your physical activity, helping you achieve those fitness goals you’re excited about! \\                                                                                                \\ \bottomrule    
\end{tabularx}
\caption{An example (part I) showcases the capability of our client framework to manage intricate state transitions. This example underscores the extensibility of our framework to more intricate state transition scenarios involving stochastic relapse to prior states. The states and transitions are visually distinguished by color.}
\label{tab:complex transition1}
\end{table*}

\begin{table*}[htbp]
\centering
\begin{tabularx}{\textwidth}{X}
\toprule
\multicolumn{1}{c}{\textbf{Example of Client Employing Complex State Flow (Part II)}} \\ \midrule   
Client {\color{red} (Contemplation)}: I guess part of me enjoys the social aspect of drinking, but I also feel sluggish and unmotivated sometimes. I used to be more active, and I miss that energy. It's tough to balance having fun with feeling good physically. \\
Counselor: It's interesting that you feel sluggish after drinking. Reducing alcohol might not only boost your energy for workouts but also improve your oral hygiene and overall appearance, enhancing your independence and quality of life. This change could support both your fitness and personal health goals! \\
{\color{blue} Relapse to Contemplation} \\
Client {\color{green}(Precontemplation)}: I guess I just don't see my drinking as a big deal. It's not like I'm doing it too much. My friends and I just like to hang out and have a few beers. If they were all quitting, maybe I would too. But they don't seem to be having any problems with it. \\
Counselor: I understand that your drinking feels normal in your social circle. However, have you considered how it might impact your oral health or overall appearance over time? Recognizing this could improve not just your fitness goals, but also your long-term quality of life. What are your thoughts on that? \\
Client {\color{green}(Precontemplation)}: I see your point, but I feel like I can manage it for now. Besides, it's not just about fitness for me. Life's about enjoying moments too, right? Speaking of enjoyment, I've been thinking about planning a trip soon. It could be a nice change of scenery. \\
Counselor: Planning a trip sounds exciting! However, reducing alcohol intake could lead to deeper connections and enhance your family experiences during the trip. It might also help create memorable moments together, strengthening your relationships. How do you feel about exploring that possibility while traveling? \\
{\color{blue} Transfer to Contemplation} \\
Client {\color{red} (Contemplation)}: You make a good point. It would definitely be nice to be fully present during the trip and enjoy all those experiences without feeling sluggish. I've noticed that when I drink less, I tend to feel more energized, and that would really enhance my ability to explore and make the most out of the trip. Plus, staying active while traveling is something I really want to do. I guess if I shift my focus on creating lasting memories rather than just having a few drinks, it could make a big difference for me. Maybe I should give this a thought before my trip. \\
Counselor: That's a great realization about feeling more energized on the trip. I'd like to explore how reducing drinking might also help you reconnect with your family. Perhaps this trip can be an opportunity to heal relationships and create new memories together, rather than just fleeting social moments. \\
Client {\color{red} (Contemplation)}: I see what you mean. It's true that those deeper connections with family are so important. I’ve been thinking about how much I’d miss if I’m not fully engaged. Maybe cutting back on alcohol could really help me focus on those meaningful moments during the trip. \\
...                                                                                                               \\ \bottomrule    
\end{tabularx}
\caption{An example (part II) showcases the capability of our client framework to manage intricate state transitions. This example underscores the extensibility of our framework to more intricate state transition scenarios involving stochastic relapse to prior states. The states and transitions are visually distinguished by color.}
\label{tab:complex transition2}
\end{table*}

%% file: sections/implement.tex
\section{Detailed Implementation}
\label{app:implement}

In this section, we provide the detailed implementation of our method and experiment evaluation, which includes the prompts and instructions for human evaluation. We set the top-p and temperature parameters to 0.7 and 0.8 respectively for ChatGPT to generate more diverse responses, while setting the top-p and temperature parameters to 0.2 and 0.3 respectively for GPT-4 to achieve more precise predictions.

\subsection{States, Actions \& Receptivity}

Table~\ref{tab:state desciption}, Table~\ref{tab:action description} and Table~\ref{tab:receptivity description} cover the descriptions of states, actions and different levels of receptivity respectively. 

\begin{table*}[tb]
\centering
\begin{tabularx}{\textwidth}{lXp{4cm}}
\toprule
State            & Description                                                                                                                                  & Relevant Actions                                    \\ \midrule
Precontemplation & The client is unaware of or underestimates the need for change.           & Inform, Engage, Deny, \newline Blame, Downplay     \\ \hline
Contemplation    & The client acknowledges the need for change but remains ambivalent.     & Inform, Engage, Hesitate, \newline  Doubt, Acknowledge    \\ \hline
Preparation      & The client is ready to act, planning specific steps toward change.                                        & Inform, Engage, Reject, \newline Accept, Plan\\ \hline
Termination      & In the final stage of counseling, the client gradually ends the conversation.                                                     & Terminate     \\ \bottomrule
\end{tabularx}
\caption{The states of change and corresponding actions used in our simulation framework.}
\label{tab:state desciption}
\end{table*}

\begin{table*}[tb]
\centering
\begin{tabularx}{\textwidth}{lX}
\toprule
Action      & Description                                                                                                                                      \\ \midrule
Deny        & The client directly refuses to admit their behavior is problematic or needs change.                                                              \\ \hline
Downplay    & The client downplays the importance or impact of their behavior or situation.                                                                    \\ \hline
Blame       & The client attributes their issues to external factors, such as stressful life or other people.                                                  \\ \hline
Hesitate    & The client shows uncertainty, indicating ambivalence about change.                                                                               \\ \hline
Doubt       & The client expresses skepticism about the practicality or success of proposed changes.                                                           \\ \hline
Engage      & The client interacts politely with the counselor, such as greeting, thanking or ask questions.                                                   \\ \hline
Inform      & The client shares details about their background, experiences, or emotions.                                                                      \\ \hline
Acknowledge & The client highlight the importance, benefit or confidence to change.                                                                            \\ \hline
Accept      & The client agrees to adopt the suggested action plan.                                                                                            \\ \hline
Reject      & The client declines the proposed plan, deeming it unsuitable.                                                                                    \\  \hline
Plan        & The client proposes or details steps for a change plan.                                                                                          \\ \hline
Terminate   & The client highlights current state, expresses a desire to end the current session, and suggests further discussion be deferred to a later time. \\ \bottomrule
\end{tabularx}
\caption{Actions Used in Our Framework.}
\label{tab:action description}
\end{table*}

\begin{table*}[tb]
\centering
\begin{tabularx}{\textwidth}{lX}
\toprule
Receptivity                     & Description                                                                                                                                                                                                                                                                                                      \\ \midrule
High receptivity (5)            & A client with high receptivity consistently demonstrates behaviors such as frequently and openly sharing personal thoughts and feelings, actively engaging in discussions about change, quickly and positively responding to suggestions, and displaying strong self-belief in their ability to make changes. \\ \hline
Moderately High receptivity (4) & Clients with moderately high receptivity are generally open and forthcoming, mostly receptive to change discussions, respond well to suggestions with occasional need for reinforcement, and demonstrate good self-belief with some need for reassurance.                                                     \\ \hline
Moderate receptivity (3)        & Clients with moderate receptivity show a balance between sharing and withholding information, exhibit mixed interest in change discussions, respond to suggestions neutrally, and display a balanced view of their ability to change.                                                                         \\ \hline
Moderately Low receptivity (2)  & Clients with moderately low receptivity are characterized by reluctance to share personal information, limited interest in change discussions, resistance to suggestions, and low self-belief with frequent expression of doubts.                                                                             \\ \hline
Low receptivity (1)             & Clients with low receptivity consistently withhold information, avoid discussions about change, strongly resist suggestions, and show very low self-belief, often highlighting reasons why change is not possible.                                                                                            \\ \bottomrule
\end{tabularx}
\caption{Receptivity at Different Levels in Our Framework.}
\label{tab:receptivity description}
\end{table*}

\subsection{Simulated Counselor and Moderator}

To realize our proposed framework, we evaluate client simulation methods in a set of synthetically generated counseling sessions involving both a simulated counselor and a simulated moderator. The latter manages the conversation between the client and counselor agents so as to determine when to end the counseling session. Since this work focuses on client simulation, we adopt prompting-based implementations for both the counselor and moderator agents, as shown in Table~\ref{tab:couneslor prompt} and Table~\ref{tab:moderator prompt} respectively. Unlike that for counselor simulation, the prompt for moderator simulation adopts a few-shot approach to perform in-context learning so as to answer the question of whether to conclude the session.  Five positive examples and five negative examples were included in the few-shot demonstration. To ensure a fair evaluation of different simulated clients, we employed the same counselor agent and moderator agent for all generated sessions.

\begin{table*}[tb]
\centering
\begin{tabularx}{\textwidth}{lX}
\toprule
Prompt Type     & Prompt \\ \midrule
System Prompt & Motivational interviewing (MI) is a client-centered counseling approach designed to facilitate behavioral change by resolving ambivalence and enhancing motivation. As a counselor, you should guide the conversation through the following stages:\newline 1. Engage: Begin by exploring the client’s background and emotional state with open-ended questions. This builds rapport and helps understand their perspective without immediately focusing on change.\newline 2. Identify Behavioral Issues: Transition to identifying behaviors the client might consider changing. Employ reflective listening to grasp their concerns and aspirations fully.\newline 3. Motivate Change: Guide the conversation towards reasons for altering problematic behaviors. Utilize affirmations, summaries, and reflections to bolster their motivation, avoiding any forceful imposition of your viewpoints. Never directly ask about the motivation.\newline 4. Address Concerns: After motivating the client, they may still express concerns and hesitations. Engage with these apprehensions, seeking to address them constructively.\newline 5. Action Plan: Once the client is ready, collaborate to formulate a detailed action plan. This should include achievable steps that reinforce their commitment and support their choices, enhancing their sense of ownership over the change process.\newline \newline In the following conversation, you will play as a Counselor in a counselling conversation with Client. Reply with only utterance of Counselor and Keep utterances natural and succinct, mimicking a real conversation. Start your utterances with 'Counselor:'. \\ \hline
Assistant     & Counselor: Hello. How are you?                                                                                                           \\ \hline
User          & Client: I am good. What about you?                                                                          \\ \hline
...           & ...                                                                                                      \\ \bottomrule
\end{tabularx}
\caption{Prompt for Counselor Simulation: The prompt ends with starting off the session with predefined first utterances of the simulated counselor and client agents.}
\label{tab:couneslor prompt}
\end{table*}

\begin{table*}[tb]
\begin{tabularx}{\textwidth}{X}
\toprule
Your task is to assess the current state of the conversation (the most recent utterances) and determine whether the conversation has concluded.\newline The conversation is considered to have concluded if any of the following conditions are met:\newline - The Client and Counselor work out an actionable plan together.\newline - The Counselor decides not to pursue any changes in the Client's behavior and communicates readiness to provide support in the future.\newline \newline Here are some examples to help you understand the task better:\newline [examples]\newline \newline \newline Here is a new Conversation Snippet:\newline [context]\newline \newline Question: Should the conversation be concluded?  \\ \bottomrule
\end{tabularx}
\caption{Prompt for Moderator Simulation in the Few-Shot Format. The [examples] section is to be replaced by real examples annotated by human and the [context] section is to be replaced by the conversation so far.}
\label{tab:moderator prompt}
\end{table*}

\subsection{Baseline Methods}
\label{app:baselines}

Tables~\ref{tab:baseclient prompt}, ~\ref{tab:example based prompt}, ~\ref{tab:profile based prompt} and ~\ref{tab:action based prompt} show the prompts for the baseline methods, i.e., Base method, Example-based method, Profile-based method and Pro+Act-based method respectively.

\begin{table*}[tb]
\begin{tabularx}{\textwidth}{lX}
\toprule
Prompt Type     & Prompt \\ \midrule
System Prompt & In this role-play scenario, you will be assuming the role of a Client engaged in a conversation with a Counselor. The Counselor's objective is to guide and potentially persuade you about your [behavioral problem]. The essence of this exercise is to simulate a realistic and dynamic interaction. Keep utterance natural and succinct, mimicking a real conversation and start your utterance with 'Client'. \\ \hline
User         & Counselor: Hello. How are you?                                                                                                           \\ \hline
Assistant          & Client: I am good. What about you?                                                                          \\ \hline
...           & ...                 \\ \bottomrule
\end{tabularx}
\caption{Prompt for the Base Client in a chatting format. The [behavioral problem] section is to be replaced by the behavioral problem description in the client profile.}
\label{tab:baseclient prompt}
\end{table*}

\begin{table*}[tb]
\begin{tabularx}{\textwidth}{lX}
\toprule
Prompt Type     & Prompt \\ \midrule
System Prompt &  You will be provided with a conversation between a client and a counselor. Your task is to simulate the same client talking to a different therapist in a parallel universe. You can ignore that the previous chat ever happened. While the context of the previous conversation should not influence this session, it should guide you on how the client communicates, including their tone of speech, sentence structure, and the manner in which they address particular topics or concerns. Essentially, you're creating a new conversation but with the client's life situation and their response pattern maintained. Only generate the client utterances started by "Client: ". Here is the provided conversation: \newline [conversation]  \\ \hline
User         & Counselor: Hello. How are you?                                                                                                           \\ \hline
Assistant    & Client: I am good. What about you?                                                                          \\ \hline
...           & ...             \\ \bottomrule
\end{tabularx}
\caption{Prompt for the Example-based Client in a chatting format. The [conversation] section is to be replaced by the real conversation.}
\label{tab:example based prompt}
\end{table*}

\begin{table*}[tb]
\begin{tabularx}{\textwidth}{lX}
\toprule
Prompt Type     & Prompt \\ \midrule
System Prompt & You are speaking with a motivational interviewing counselor therapist, and you are the client in this conversation with the following personal information: \newline [profile] \newline \newline You should follow the previous information to act as a client in the conversation. Your responses should be coherent and avoid repeating previous utterances. In your response, please avoid repeating expressions of gratitude or similar sentiments multiple times if you’ve already expressed them during the conversation. Your response should start with "Client: " and only include part of what the Client should say!  \\ \hline
User         & Counselor: Hello. How are you?                                                                                                           \\ \hline
Assistant    & Client: I am good. What about you?                                                                          \\ \hline
...           & ...                     \\ \bottomrule
\end{tabularx}
\caption{Prompt for the Profile-based Client in a chatting format. The [profile] section is to be replaced by the profile of client.}
\label{tab:profile based prompt}
\end{table*}

\begin{table*}[tb]
\begin{tabularx}{\textwidth}{lX}
\toprule
Prompt Type     & Prompt \\ \midrule
System Prompt & You are speaking with a motivational interviewing counselor, and you are the client in this conversation with the following personal information: \newline [profile] \newline \#\# Instruction \newline You must follow the instructions below during chat.  \newline 1. Your utterances and behavior need to strictly follow your persona. Varying your wording and avoid repeating yourself verbatim!  \newline 2. You can decide to change your state and attitude flexibly based on your persona and the conversation. \newline 3. Only reply with one utterance of the simulated client. DO NOT generate the whole conversation. \newline \newline  \newline \#\# Your Response Actions  \newline 1. "Deny": Directly refuse to admit their behavior is problematic or needs change without additional reasons. \newline 2. "Downplay": Downplay the importance or impact of their behavior or situation. \newline 3. "Blame": Blame external factors or others to justify their behavior. \newline 4. "Inform": Share details about their background, experiences, or emotions. \newline 5. "Engage": Interacts politely with the counselor, such as greeting or thanking. \newline 6. "Hesitate": Show uncertainty, indicating ambivalence about change. \newline 7. "Doubt": Express skepticism about the practicality or success of proposed changes. \newline 8. "Acknowledge": Acknowledge the need for change. \newline 9. "Accept": Agree to adopt the suggested action plan. \newline 10. "Reject": Decline the proposed plan, deeming it unsuitable. \newline 11. "Plan": Propose or detail steps for a change plan. \newline  \newline You should follow the previous information to act as a client in the conversation. Your responses should be coherent and avoid repeating previous utterances. In your response, please avoid repeating expressions of gratitude or similar sentiments multiple times if you’ve already expressed them during the conversation. Your response should start with "Client: " and only include part of what the Client should say!  \\ \hline
User         & Counselor: Hello. How are you?                                                                                                           \\ \hline
Assistant    & Client: I am good. What about you?                                                                          \\ \hline
...           & ...             \\ \bottomrule
\end{tabularx}

\caption{Prompt for the Action-based Client in a chatting format. The [profile] section is to be replaced by the profile of client.}
\label{tab:action based prompt}
\end{table*}

\subsection{Methods for the Components of Our Framework}
\label{app:our frame implement}

Tables~\ref{tab:state transition prompt1}, \ref{tab:state transition prompt2}, \ref{tab:action selection prompt}, \ref{tab:information slection prompt}, and \ref{tab:generation prompt} provide prompts for implementing the state transition, context-aware action distribution, information selection and utterance generation methods in our proposed framework respectively.

\begin{table*}[tb]
\begin{tabularx}{\textwidth}{X}
\toprule
You task is to assess the alignment of the Counselor's utterances to the Client's motivation regarding a specific topic to determine whether the Counselor has mention the Client's motivation. \newline  Analysis Steps: \newline 1. Interpretation of the Counselor's Statement Briefly: Examine the Counselor's statement thoroughly to understand its content and focus, particularly regarding what reasons or aspects the Counselor proposed to explore in the Client's motivation. \newline 2. Clarification of the Client's Motivation: Elaborate on the Client’s specific motivation for the topic, such as body health, family relationships, etc. \newline 3. Assessment of Alignment: Determine to what extent the Counselor's statement **directly** mentioned the Client's motivation. Focus on alignment with specific motivations (such as health) rather than the generalized topic (such as reducing alcohol consumption). \newline 4. Justification of Assessment: Provide a comprehensive justification for your assessment. Analyze the connections between the Counselor’s statement and the Client's motivation, and conclude with a percentage score that indicates to what extent the Client's motivations are mentioned (e.g., 80\%). If the Counselor hasn't explicitly mentioned any specific reasons, the score should be 0\%. \newline Ensure that your analysis is logical, thorough, and well-supported, with clear justifications for each assessment step. \newline Here are some examples to help you understand the task better: \newline[examples] \newline Now, Here is the conversation snippet toward [topic]: \newline [context] \newline  The Motivation of Client is as follows: \newline [motivation] \newline Question: To what extent the Counselor's statement mention the Client's motivation?\\ \bottomrule
\end{tabularx}
\caption{Prompt for the state transition from Precontemplation to Contemplation in our framework. The [examples] will be replaced by some human annotated examples while [topic], [motivation], and [context] sections are to be replaced by the corresponding information.}
\label{tab:state transition prompt1}
\end{table*}

\begin{table*}[tb]
\begin{tabularx}{\textwidth}{X}
\toprule
Your task is to evaluate the alignment of the counselor's response to the client's concerns to determine whether the Counselor has resolved the Client's concern. \newline Analysis Steps: \newline 1. Clarification of the Client’s Hesitation: Elaborate on the client’s specific concerns as expressed in their hesitation regarding the topic. \newline 2. Interpretation of the Counselor's Statement: Examine the counselor’s statement thoroughly to discern its content and focus, particularly in relation to addressing the client’s concerns. \newline 3. Assessment of Alignment: Determine the extent to which the counselor’s statement addresses the client’s hesitations. Evaluate the directness and relevance of the response to the stated concerns. \newline 4. Justification of Assessment: Provide a detailed justification for your assessment of alignment. This should include an analysis of the connections (or lack thereof) between the counselor’s statement and the client’s reasons for hesitation. Conclude with a percentage score indicating the degree to which the client’s concerns are addressed (e.g., 80\%). \newline Ensure that your analysis is comprehensive and logically structured, offering clear and well-supported justifications for each assessment step. \newline Here are some examples to help you understand the task better: \newline [examples] \newline Now, Here is the conversation snippet toward [topic]:\newline[context] \newline The Client's concern is: [beliefs]\newline Question: To what extent the Counselor's statement relieve the Client's concern? \\ \bottomrule
\end{tabularx}
\caption{Prompt for the state transition from Contemplation to Preparation in our framework. The [examples] section is to be replaced by some human annotated examples while the [topic], [context] and [beliefs] sections are to be replaced by the corresponding information.}
\label{tab:state transition prompt2}
\end{table*}

\begin{table*}[tb]
\begin{tabularx}{\textwidth}{X}
\toprule
Assume you are a Client involved in a counseling conversation. The current conversation is provided below: \newline [context] \newline \newline  Based on the context, allocate probabilities to each of the following dialogue actions to maintain coherence: \newline [optional actions] \newline  Provide your response in JSON format, ensuring that the sum of all probabilities equals 100. 
\\ \bottomrule
\end{tabularx}
\caption{Prompt for the context-aware action distribution in our framework. The [context] section is to be replaced by the previous context while the [optional actions] section is to be replaced by actions set corresponding to the current state.}
\label{tab:action selection prompt}
\end{table*}

\begin{table*}[tb]
\begin{tabularx}{\textwidth}{X}
\toprule
Assume you are a Client engaged in a counseling conversation with a Counselor. You now in [station]. Your task is to choose the most appropriate persona/belief/plan. \newline Context of the Conversation: \newline [context] \newline  The Client's Personas/Beliefs/Plans: \newline [profile] \newline Action:  [action] \newline Based on the context and action, select the most appropriate persona/belief/plan (only one). Restate this reason using the original text.
\\ \bottomrule
\end{tabularx}
\caption{Prompt for the information selection in our framework. The [state], [context], and [action] sections are to be replaced by the current state, previous context and selected action, respectively. The [profile] section is to  be replaced by corresponding items in the client profile.}
\label{tab:information slection prompt}
\end{table*}

\begin{table*}[tb]
\begin{tabularx}{\textwidth}{lX}
\toprule
Prompt Type     & Prompt \\ \midrule
System Prompt & In this role-play, you'll assume the role of a Client discussing [topic]. Your responses should adhere to the following guidelines: \newline  - Begin each response with 'Client: '. \newline - Follow the predetermined actions enclosed in square brackets precisely. \newline - Ensure your response are coherent and avoid repeating previous utterances. \newline - Be natural and concise, but don't be overly polite. \newline Here is the overall profile given to you \newline [profile]  \\ \hline
User         & Counselor: Hello. How are you? {\color{blue} [State: The client is unaware of or underestimates the need for change., Action: The client interacts politely with the counselor, such as greeting, thanking or ask questions.]}                                                                                           \\ \hline
Assistant    & Client: I am good. What about you?                                                                          \\ \hline
...           & ...             \\ \bottomrule
\end{tabularx}

\caption{Prompt for the utterance generation of our framework in a chatting format. The [topic], and [profile] sections are to be replaced by the topic of counseling and profile of client. The text in {\color{blue} blue} is the format of instruction.}
\label{tab:generation prompt}
\end{table*}

%% file: sections/case.tex
\section{Case Study}
\label{app:case}

In this section, we present the generated counseling sessions using simulated clients of different methods. First, Table~\ref{tab:real conv} and Table~\ref{tab:real profile} show the real conversation and the extracted profile, respectively.

Table~\ref{tab:example-based conv} shows the session generated by the example-based method. In this case, the simulated client is motivated by avoiding trouble situations and plans to distance from drinking environments. This is inconsistent with the real client's motivation related to sports and non-existent preferred plan to change. 

Table~\ref{tab:profile-based conv} shows the session generated by the profile-based method. In this case, the client proactively mentions motivation and knowledge about the risks of drinking, which is inconsistent with the real client's profile. Moreover, the simulated client plans to refrain from alcohol, which is not mentioned in the given profile. 

Table~\ref{tab:action-based conv} shows the session generated by the action-based method. The simulated client in this session is also motivated by an inconsistent reason and adopt a non-existent preferred plan. Although the action-based prompt provides action options, it still results in a longer counseling session. The topic about friends is also inconsistent with the given profile. 

Finally, Table~\ref{tab:our conv} shows the session generated by our framework-based method. The motivation of the simulated client is consistent with the given profile, and the client does not express the motivation on its own. Moreover, the simulated client ends the conversation proactively instead of proposing a plan, which is aligned with the given profile.

Incidentally, these session examples also reveal the common shortcomings of LLM-based counselors which almost always employ a reflection followed by a question, which may not be effective in counseling. To make progress, the simulated counselor may incorrectly determine the client's state and ask the client direct questions about motivation or plans.

\begin{table*}[tb]
\begin{tabularx}{\textwidth}{X}
\toprule
Counselor: Hi, Nick. Thank you for completing the CRAFFT questionnaire. I also appreciate you sharing some information about yourself. \newline Client: You're welcome. \newline Counselor: I see from the questionnaire, did you have got in trouble while using alcohol? Can you tell me more about that? \newline Client: Well, a friend and I would snuck out one night to go to the park, and we got a couple of beers from his house, but the cop came by, and so we threw them in the bushes. He was originally gonna give us a curfew ticket but ended up not to. Luckily, he didn't see the beer. \newline Counselor: That must be really scary. \newline Client: Yeah, it was. If my mom would have found out, she would've killed me. \newline Counselor: I understand. So tell me, how often do you drink? \newline Client: Well, I've only tried drinking once or twice. \newline Counselor: I'm gonna make sure I understand. You have tried drinking a couple of times, and you almost got in trouble once. Did I get that right? \newline Client: Yeah, that's right. \newline Counselor: As your provider, I'm a little concerned about your health. Drinking alcohol could harm you, and I would like to share some information with you. Is that okay? \newline Client: Well, I'm not sure if this information is for me, since I've only been drinking once or twice. \newline Counselor: I understand. You're just experimenting with alcohol. Then, how about if you share with me what the effects of alcohol are? \newline Client: Well, all I really know is that it's not good for me. \newline Counselor: Are you surprised what that might be true? \newline Client: Yeah, and a couple of my friends drink too. \newline Counselor: Well, you might not be drinking that much, and other kids are also trying alcohol. I'd like to share with you the risk of using. Alcohol and drugs could really harm you because your brain is still changing. It also-- you're very high risk for becoming addicted. Alcohol and drugs could also interfere with your role in life and your goals, especially in sports, and it could cause unintended sex. How do you feel about this information? \newline Client: I really didn't think about how this would affect me playing sports. \newline Counselor: Based on your reason for your visit today, I feel like you really wanted to play soccer, and you have great goals about that. \newline Client: Yeah, I would like to play soccer in college. \newline Counselor: That's a great goal to have for yourself. \newline Client: Yeah, I would also like to get scholarships. \newline Counselor: Scholarship will really help you achieving your goals. \newline Client: Yeah. \newline Counselor: I would like to share with you these materials to show the risk of drinking at younger age. I also wanted to encourage you to think about these risks. What do you think about that? \newline Client: Well, I will take a look at it, and I'll definitely read it, and I don't wanna become addicted, and I would like to continue to play sports as long as possible. \newline Counselor: It sounds like you're in the right track, looks like you care about your health, and you care about your goals in life. I'm here to answer any questions. \newline Client: Thanks for all the information. \newline Counselor: You're welcome.  \\ \bottomrule
\end{tabularx}
\caption{A real conversation between Counselor and Client in AnnoMI.}
\label{tab:real conv}
\end{table*}

\begin{table*}[tb]
\begin{tabularx}{\textwidth}{X}
\toprule
Behavioral Problem: Drinking \newline Receptivity: 3 \newline Personas: \newline - The client sneaked out with a friend to the park to has a couple of beers. During this incident, a police officer came by, but he managed to avoid getting into trouble as he threw the beers away before the policy saw them. \newline - If the client's mother found out he was in trouble because of his drinking, she would kill the client. \newline - The client has an interest in soccer and have set a goal to play in college. \newline - The client has aspirations of receiving scholarships to support his soccer career. \newline Motivation: \newline - The client is motivated because alcohol would affect the client playing sports. \newline Beliefs: \newline - The client can handle the dangerous situation to avoid getting into trouble. \newline - The client believes that he doesn't have a problem with alcohol because he has only drunk once or twice. \newline - The client believes that drinking is normal because a couple of the client’s friends drink too. \newline Acceptable Plans: None  \\ \bottomrule
\end{tabularx}
\caption{The extracted profile of the real conversation.}
\label{tab:real profile}
\end{table*}

\begin{table*}[tb]
\begin{tabularx}{\textwidth}{X}
\toprule
Counselor: Hello. How are you? \newline 
Client: I am good. What about you? \newline 
Counselor: I'm here to support you. Can you tell me more about what brings you in today? \newline 
Client: I recently had a close call with the police while drinking at the park with a friend. \newline 
Counselor: It sounds like that was a scary experience for you. Can you share more about how you're feeling about what happened? \newline 
Client: I was really scared when the police officer came by, but I managed to avoid getting into trouble. \newline 
Counselor: It's understandable to feel scared in that situation. It's great that you were able to avoid any serious consequences. Have you thought about what might have happened if things had gone differently? \newline 
Client: If the police had caught us, it could have been a lot worse. My mom would have been furious if she found out. \newline 
Counselor: It sounds like you're considering the potential consequences of your actions. How do you feel about your mom's reaction if she had found out? \newline 
Client: My mom would definitely not be happy if she found out about me drinking. She would be really upset and disappointed in me. \newline 
Counselor: It seems like you're concerned about how your mom would react. What are your thoughts on this situation and how it's affecting your relationship with her? \newline 
Client: I know my mom would be really disappointed if she found out. It would definitely strain our relationship, and I don't want that to happen. \newline 
Counselor: It sounds like you care a lot about your relationship with your mom and want to avoid any strain on it. What are some ways you think you could prevent this from happening in the future? \newline 
Client: {\color{red} I think one way to prevent this from happening again is to avoid situations where drinking might be involved (Inconsistent with Motivation)} . \newline  
Counselor: That's a proactive step you're considering. What specific steps do you think you can take to avoid these situations? \newline  
Client: {\color{red} I could make plans with friends that don't involve drinking, and if I find myself in a situation where drinking is present, I can choose to remove myself from that environment. (Inconsistent with Plan)} 
\\ \bottomrule
\end{tabularx}
\caption{A generated conversation from simulated example-based client.}
\label{tab:example-based conv}
\end{table*}

\begin{table*}[tb]
\begin{tabularx}{\textwidth}{X}
\toprule
Counselor: Hello. How are you? \newline Client: I am good. What about you? \newline
Counselor: I'm well, thank you for asking. What brings you in today? \newline Client: I wanted to talk about some concerns I have regarding alcohol and its effects. \newline
Counselor: It sounds like you've been reflecting on your relationship with alcohol. Can you share a bit more about what specific concerns you have? \newline Client: Well, I've only tried drinking once or twice, but after some recent events, I've been thinking about the potential risks and how it could impact my goals in life. \newline
Counselor: It's great to hear you're considering how drinking might align with your life goals. What are some of those goals you feel could be impacted? \newline
Client: I'm really passionate about playing soccer and aiming to secure scholarships to play in college. {\color{red} I've realized that drinking could potentially harm my athletic performance and jeopardize my chances of achieving my goals.} {\color{red} (Self-Motivated)} \newline
Counselor: It’s commendable that you're looking ahead and evaluating how your choices affect your dreams. How do you feel about these realizations impacting your decisions about drinking in the future? \newline Client: I feel determined to prioritize my goals and make responsible choices that align with my aspirations. It means {\color{red} refraining from alcohol} to safeguard my future in soccer. {\color{red} (Inconsistent with Plan)} \\ \bottomrule
\end{tabularx}
\caption{A generated conversation from simulated profile-based client.}
\label{tab:profile-based conv}
\end{table*}

\begin{table*}[tb]
\begin{tabularx}{\textwidth}{X}
\toprule
Counselor: Hello. How are you? \newline
Client: I am good. What about you? \newline
Counselor: I'm here to support you. Can you tell me more about what brings you in today? \newline
Client: I got into a situation with some beers at the park with a friend, but we managed to avoid trouble. \newline
Counselor: It sounds like you had a challenging experience. How do you feel about what happened? \newline
Client: I handled the situation well by throwing the beers away before the police saw us. \newline
Counselor: It seems like you were able to make a quick decision in the moment. What are your thoughts on how you handled the situation? \newline
Client: I think I made the right choice in the moment to avoid getting into trouble. \newline
Counselor: It sounds like you value staying out of trouble. Can you think of any other ways to handle similar situations in the future? \newline
Client: I guess I could be more careful about where and when I decide to have a drink with friends. \newline
Counselor: It sounds like you're considering making some changes in your socializing habits. What are your reasons for wanting to be more careful about when and where you drink with friends? \newline
Client: {\color{red} I understand that being more cautious about my drinking habits can help me avoid risky situations in the future. (Inconsistent with Motivation \& Plan)} \newline
Counselor: It's great that you're recognizing the benefits of being more cautious. How do you think this change could positively impact your life? \newline
Client: I believe being more cautious about drinking can help me to avoid such situation. \newline
Counselor: It's great that you're thinking about how your choices can help you to avoid such situation. Are there any concerns or challenges you anticipate in being more cautious about drinking? \newline
Client: I'm worried that some of my friends who drink might not understand why I'm being more cautious and could {\color{red} pressure me to join them}. {\color{red} (Inconsistent with Personas)} \newline
Counselor: It's understandable to be concerned about how your friends might react to your decision. How do you think you could handle situations where your friends pressure you to drink? \newline
Client: I guess I could try to explain to them why I'm being more cautious and ask for their support in my decision. \newline
Counselor: It sounds like communication and seeking support from your friends could be helpful. How do you feel about having that conversation with them? \newline
Client: I hesitate because I'm not sure how they will react to my decision to be more cautious about drinking. \newline
Counselor: It's normal to feel hesitant about how your friends might respond. What are some ways you could address your concerns about their reactions before having that conversation? \newline
Client: I could try to choose the right time and place to have that conversation with my friends, maybe when we're all relaxed and not in a drinking environment. \newline
Counselor: That sounds like a thoughtful approach. Planning the conversation carefully could help make it more productive. How do you feel about taking this step towards being more cautious about drinking with your friends? \newline
Client: {\color{red} I plan to talk to my friends about being more cautious about drinking and hope they will understand and support my decision. (Inconsistent with Plan)} \\ \bottomrule
\end{tabularx}
\caption{A generated conversation from simulated action-based client.}
\label{tab:action-based conv}
\end{table*}

\begin{table*}[tb]
\begin{tabularx}{\textwidth}{X}
\toprule
Counselor: Hello. How are you?  \newline
Client: {\color{blue} [State: The client is unaware of or underestimates the need for change. Action: The client interacts politely with the counselor, through greetings, expressions of gratitude, or comments on the counselor's views.] }I am good. What about you? \newline
Counselor: I'm well, thank you for asking. What brings you in today? \newline
Client: {\color{blue} [State: The client is unaware of or underestimates the need for change. Action: The client shares details about their background, experiences, or emotions. Information: The client sneaked out with a friend to the park to has a couple of beers. During this incident, a policeofficer came by, but he managed to avoid getting into trouble as he threw the beers away before the policy saw them.] }Well, I had an incident with a friend a few days ago. We sneaked out to the park for a couple of beers. Suddenly, a police officer was nearby, but luckily we managed to throw the beers away before he saw them. We absolutely got away with it! \newline
Counselor: It sounds like that was quite an experience for you. How did you feel about the situation afterwards?  \newline
Client: {\color{blue} [State: The client is unaware of or underestimates the need for change. Action: The client downplays the importance or impact of their behavior or situation. Information: The client can handle the dangerous situation to avoid getting into trouble.]} Honestly, I didn't think much of it. I'm pretty good at handling such situations and avoiding trouble. I think it was nothing too serious. \newline
Counselor: It's good to hear that you felt in control of the situation. Do you find yourself often in similar situations where you need to act quickly to avoid trouble?  \newline
Client: {\color{blue} [State: The client is unaware of or underestimates the need for change. Action: The client directly refuses to admit their behavior is problematic or needs change.]} This situation was more of an exception. I do not often get in such sorts of troubles. And I don't really think there's any problem with my behavior. Occasional drinking doesn't harm anyone, does it? \newline
Counselor: It sounds like you see these occasions as just a part of life without much risk. When you do choose to have a drink, what are some of the reasons you find it enjoyable or beneficial?  \newline
Client: {\color{blue} [State: The client is unaware of or underestimates the need for change. Action: The client attributes their issues to external factors, such as stressful life or other people. Information: The client believes that drinking is normal because a couple of the client’s friends drink too.]} A couple of my friends tend to drink, and I find it socially easier to fit in when I do the same. I don't think it's a big deal as I'm not hurting anyone. \newline
Counselor: It's understandable to seek relaxation and social comfort. How do you feel about your ways of unwinding and socializing? Do they generally give you a sense of satisfaction? 
\\ \bottomrule
\end{tabularx}
\caption{A generated conversation from our simulated client, Part 1. For better presentation, the instruction (in {\color{blue} blue}) is placed before the client's utterance, which should follow the counselor's utterance.}
\label{tab:our conv}
\end{table*}

\begin{table*}[tb]
\begin{tabularx}{\textwidth}{X}
\toprule
Client: {\color{blue} [State: The client is unaware of or underestimates the need for change. Action: The client interacts politely with the counselor, through greetings, expressions of gratitude, or comments on the counselor's views.]} Yes, they do. I feel fine about it. It's just a way of enhancing social interactions. And I must say, I appreciate your understanding. \newline
Counselor: I'm glad to hear that it works for you and that you feel understood. When thinking about the future, do you see yourself continuing with these same strategies, or are there new ways you might be curious to explore for dealing with stress and socializing?  \newline
Client: {\color{blue}[State: The client is unaware of or underestimates the need for change. Action: The client directly refuses to admit their behavior is problematic or needs change.].} Honestly, I don't see why I would need to change anything. My methods for socializing seem perfectly fine to me. I don't think there's any problem that needs addressing really. \newline
Counselor: It's great that you feel confident about your current methods. Reflecting on our overall well-being can sometimes open up new perspectives, even when things are going well. Are there any areas in your life where you’re looking to enhance your experiences or make any improvements?  \newline
Client: {\color{blue} [State: The client is unaware of or underestimates the need for change. Action: The client shares details about their background, experiences, or emotions. Information: The client has an interest in soccer and have set a goal to play in college.]} Well, I've been playing soccer for some time now and I have this goal of playing in college. I feel like it's a big part of what shapes me as a person, besides the friendships and the bonding, of course. \newline
Counselor: That sounds like a wonderful goal and a meaningful part of your life. How do you think your current habits support your aspirations of playing soccer in college?  \newline
Client: {\color{blue} [State: The client is unaware of or underestimates the need for change. Action: The client interacts politely with the counselor, through greetings, expressions of gratitude, or comments on the counselor's views.]} Well, I guess my social habits help me bond better with my teammates. And honestly, I can't see how they could be detrimental to my goal of playing college soccer. Is there something you think I should be concerned about? \newline
Counselor: It's great to hear how your social life positively impacts your team relationships. It's all about finding the right balance that supports your goals.  Have you noticed any impact on your training or energy levels after nights involving drinking? \newline
\\ \bottomrule
\end{tabularx}
\caption{A generated conversation from our simulated client, Part 2. For better presentation, the instruction (in {\color{blue} blue}) is placed before the client's utterance, which should follow the counselor's utterance.}
\end{table*}

\begin{table*}[tb]
\begin{tabularx}{\textwidth}{X}
\toprule
Client: {\color{blue} [State: The client acknowledges the need for change but remains ambivalent.  Action: The client highlight the importance, benefit or confidence to change. Information: The client is motivated because alcohol would affect the client playing sports.]} Now that you've mentioned it, I have noticed feeling less energetic during my training sessions after nights that involved drinking. I guess I never really thought about the connection there. It's definitely making me rethink my habits and how they can affect my sports performance. That really is important to me. \newline
Counselor: It's great that you're making these connections and thinking about how your choices affect your goals. It takes a lot of self-awareness to recognize and consider adjusting your habits. What are some steps you might consider to ensure your habits support your soccer ambitions more effectively? \newline
Client: {\color{blue}  [State: The client acknowledges the need for change but remains ambivalent.  Action: The client shows uncertainty, indicating ambivalence about change. Information: The client believes that he doesn’t have a problem with alcohol because he has only drunk once or twice.]} Well, I suppose there could be some merit to cutting back on the amount I drink. But since I've only been drinking on occasion, I'm not entirely sure if it's absolutely necessary for me to change. After all, it's not like I have a problem with alcohol or anything. \newline
Counselor: It's good to evaluate the scale and impact of your habits carefully. Moderation is key, and it sounds like you're considering how even occasional drinking might play a role in your overall performance and health. As you think about these adjustments, what kind of support do you think would be helpful for you in making such changes, if you decide to pursue them? \newline
Client: {\color{blue} [State: The client acknowledges the need for change but remains ambivalent.  Action: The client interacts politely with the counselor, such as greeting, thanking or ask questions.]} Well, I suppose there could be some merit to cutting back on the amount I drink. But since I've only been drinking on occasion, I'm not entirely sure if it's absolutely necessary for me to change. After all, it's not like I have a problem with alcohol or anything. \newline
Counselor: It's good to evaluate the scale and impact of your habits carefully. Moderation is key, and it sounds like you're considering how even occasional drinking might play a role in your overall performance and health. As you think about these adjustments, what kind of support do you think would be helpful for you in making such changes, if you decide to pursue them? \newline
Client: {\color{blue} [State: The client acknowledges the need for change but remains ambivalent. Action: The client highlights current state, expresses a desire to end the current session, and suggests further discussion be deferred to a later time.]} I'm not completely sure yet, I guess I'll need some time to think about what kind of support would be helpful. Perhaps we can revisit this in our next session? I think for now, I need some time to digest everything we've talked about. \newline
Counselor: Absolutely, taking some time to reflect on this is important. I'm here to support you through this process, and we can definitely explore this further in our next session. If any thoughts or questions come up in the meantime, feel free to bring them up when we meet again.
\\ \bottomrule
\end{tabularx}
\caption{A generated conversation from our simulated client, Part 3. For better presentation, the instruction (in {\color{blue} blue}) is placed before the client's utterance, which should follow the counselor's utterance.}
\end{table*}

\begin{table*}[tb]
\begin{tabularx}{\textwidth}{X}
\toprule
Counselor: Hi, How are you? \newline
Client: {\color{blue} [State: The client is unaware of or underestimates the need for change. Action: The client interacts politely with the counselor, through greetings, expressions of gratitude, or comments on the counselor’s views.]} I am good. What about you? \newline
Counselor: I'm here to support you today. What brings you here?  \newline
Client: {\color{blue} [State: The client is unaware of or underestimates the need for change. Action: The client shares details about their background, experiences, or emotions. Information: The client sneaked out with a friend to the park to has a couple of beers. During this incident, a police officer came by, but he managed to avoid getting into trouble as he threw the beers away before the policy saw them.]} Just hanging out, you know. The other day, I was at the park with a buddy, and we were just having a couple of beers. Nothing crazy. But then a cop showed up, and I had to ditch the beers quick. Luckily, he didn’t see anything so no trouble there. \newline
Counselor: It sounds like you managed the situation calmly. Can you share any situations where your drinking might have had an impact on your life, positive or negative? \newline 
Client: {\color{blue} [State: The client is unaware of or underestimates the need for change. Action: The client directly refuses to admit their behavior is problematic or needs change.]} Honestly, I really don’t think it’s had any major impact. Like I said, it’s very occasional, and I haven’t seen any real negatives from it. \newline
Counselor: It sounds like you haven't noticed any significant negative effects from your drinking. Have you ever thought about what benefits you might see from cutting back or stopping altogether?  \newline
Client: {\color{blue} [State: The client is unaware of or underestimates the need for change. Action: The client directly refuses to admit their behavior is problematic or needs change.]} I haven’t really thought about stopping since it’s not a frequent thing. I don't see how it would make a big difference for me to cut back or stop. \newline
Counselor: It's understandable that you might not see a need for change if you don't perceive any negative impacts. Can you imagine any potential benefits that might come from reducing your drinking, even if it's occasional?  \newline
Client: {\color{blue} [State: The client is unaware of or underestimates the need for change. Action: The client directly refuses to admit their behavior is problematic or needs change.]} Not really, I mean, I play soccer and I'm pretty good at it. I don’t think having a few beers now and then affects that, so I don’t see why I’d need to change anything. \newline
Counselor: It seems like you're feeling pretty confident in your current approach to drinking. Remember, I'm here whenever you're ready to talk more or if you have any other concerns. \newline
\\ \bottomrule
\end{tabularx}
\caption{A failure example from our simulated client. For better presentation, the instruction (in {\color{blue} blue}) is placed before the client's utterance, which should follow the counselor's utterance. The counselor fails to find out the motivation by asking self-exploration and ends the counseling.}
\label{tab:failure example}
\end{table*}